\documentclass{article}

\usepackage[margin=1in]{geometry}
\usepackage{microtype}
\usepackage{graphicx}
\usepackage{subcaption}
\usepackage{booktabs}
\usepackage[numbers]{natbib}
\usepackage{hyperref}

\usepackage{amsmath}
\usepackage{amssymb}
\usepackage{mathtools}
\usepackage{amsthm}

\usepackage[capitalize,noabbrev]{cleveref}

\theoremstyle{plain}

\theoremstyle{definition}

\theoremstyle{remark}

\usepackage[disable,textsize=tiny]{todonotes}

\usepackage{url}
\usepackage{enumitem}
\usepackage{multirow}
\usepackage{makecell}
\usepackage{pifont}
\usepackage{threeparttable}
\usepackage{wrapfig}
\usepackage[utf8]{inputenc}
\usepackage[most]{tcolorbox}
\usepackage{listings}

\def\framework{Euclid-Omni}
\def\solver{Euclidea}
\newcommand{\smallsec}[1]{\textbf{#1.}}

\title{\framework{}: A Unified Neuro-Symbolic Framework \\for Plane Geometry}

\author{Zhaoyu Li$^2$\thanks{Equal contribution.
This work was partially done during Zhaoyu's internship at Meta FAIR and Kaiyu Yang's employment at Meta FAIR.
},
Hangrui Bi$^{2}$\footnotemark[1],
Youyuan Zhang$^2$,
Wenjie Ma$^3$ \\
Zenan Li$^4$,
Zhaolei Zhang$^2$,
Xujie Si$^2$,
Kaiyu Yang$^1$ \\
$^1$ Apodex, $^2$ University of Toronto, $^3$ UC Berkeley, $^4$ ETH Z\"urich \\
\texttt{zhaoyu@cs.toronto.edu}, \texttt{hangruibi@outlook.com}, \\
\texttt{windsey@berkeley.edu}, \texttt{zenan.li@inf.ethz.ch}, \\
\texttt{zhaolei.zhang@utoronto.ca}, \\
\texttt{six@cs.toronto.edu}, \texttt{kaiyu@apodex.com} \\
}

\begin{document}

\maketitle

\begin{abstract}
Euclidean geometry is a compelling testbed for AI reasoning, as it demands the combination of intuitive diagram understanding, axiomatic deduction, and algebraic computation.
Yet, existing approaches typically address only a subset of these abilities or struggle with competition-level problems.
We introduce \textit{Euclid-Omni}, a unified neuro-symbolic framework that couples a formal geometry system with Large Language Models (LLMs) and Vision-Language Models (VLMs) to tackle both calculation- and proving-style problems, in formal and natural languages, up to Olympiad-level difficulty.
At its core, we develop \textit{Euclidea}, a versatile symbolic geometry solver that automatically generates reasoning steps through deductive inference and algebraic computation.
Building on this, we develop a data-generation pipeline that synthesizes symbolic problems and solutions, renders diagrams, and translates them into natural language, producing large-scale, diverse datasets for training LLMs and VLMs across a wide range of reasoning settings.
Experiments show that VLMs trained on our synthetic data achieve superior performance on calculation tasks, and that LLMs combined with \textit{Euclidea} are competitive with state-of-the-art systems on Olympiad-level proving problems, despite using orders of magnitude less compute and training data.
Code and scripts are publicly available at \url{https://github.com/20171130/Euclid-Omni}.
\end{abstract}

\section{Introduction}
Plane geometry has been a cornerstone of mathematical education for over two millennia, ever since Euclid's \emph{Elements}, with problems ranging from elementary classroom exercises to International Mathematical Olympiad~(IMO) challenges. It also occupies a special place in the history of artificial intelligence, where it inspired some of the earliest symbolic systems for automated theorem proving~\citep{survey}. With the recent surge of large language models (LLMs) and vision-language models (VLMs), plane geometry has once again become a popular testbed for probing and improving machine reasoning~\citep{alphageometry,mathvista,mathverse}, and state-of-the-art systems~\citep{tonggeometry,alphageometry2,chen2025seed} have now reached the level of IMO gold medalists, marking a new milestone in automated geometric reasoning.

These advances, however, remain limited in both scope and accessibility.
On the scope side, IMO-level systems are tailored almost exclusively to competition-style theorem proving and offer little support for the algebraic computation that calculation-based problems require, even though such problems are equally common in plane geometry. They also tend to produce solutions that are far from how humans reason and rarely engage with natural language or visual diagrams, which restricts their usefulness in educational settings. 
The complementary line of work that targets calculation tasks faces the opposite problem: it covers only a narrow set of theorems and stops well short of competition-level difficulty~\citep{geoqa,inter-gps,unigeo}.
On the accessibility side, training IMO-level systems consumes enormous compute, and to date none has released its data-generation pipeline or training data. Public datasets, in turn, remain small, lack diversity, and are poorly stratified by difficulty~\citep{geoqa,inter-gps,geoqa_plus,unigeo}, which makes them inadequate for training modern LLMs and VLMs at scale.
Taken together, these gaps make existing approaches difficult to build on as a general foundation for research in geometric reasoning.

To bridge this gap, we introduce \textit{\framework{}}, a unified neuro-symbolic framework that targets a broad range of geometric reasoning settings.
At the heart of \framework{} is \textit{\solver{}}, a symbolic engine that exhaustively applies admissible inference rules until the deductive closure is reached.
Most existing IMO-level systems are built on full-angle notation~\citep{full-angle}, which is convenient for certain competition-style proofs.
However, it cannot distinguish an angle or arc from its supplement and is therefore ill-suited to calculation problems.
\solver{} instead grounds its axiomatic foundation directly in Euclid's \emph{Elements}~\citep{e}. This allows it to handle both calculation and proving tasks within a single framework, and yields reasoning steps that closely resemble those taught in school geometry.
We evaluate \solver{} on Geometry3K~\citep{inter-gps}, JGEX-AG-231~\citep{alphageometry}, and IMO-AG-30~\citep{alphageometry}, which together cover both calculation and proving, and find that it solves more problems than existing formal geometry systems while producing notably more human-readable solutions.

Built on top of \solver{}, our second contribution is a versatile data-generation pipeline for training LLMs and VLMs.
The pipeline produces geometry problems across text and vision modalities, in formal and natural language, and for both calculation and proving tasks. It is highly configurable, so the resulting datasets can be tailored to specific training objectives and difficulty levels, from elementary problems to IMO-level challenges.
To assess the pipeline, we use its synthetic data to train models for two settings that sit at opposite ends of the geometric reasoning spectrum: (i) multimodal geometric calculation with diagram visualizations and natural-language descriptions, and (ii) Olympiad-level formal theorem proving.
In the first setting, we train VLMs on our synthetic data and evaluate them on GeoQA~\citep{geoqa}, Geometry3K~\citep{inter-gps}, MathVista~\citep{mathvista}, and MathVerse~\citep{mathverse}; despite using substantially less training data than prior work, our models match or surpass existing approaches across these benchmarks.
In the second setting, we evaluate an LLM trained on our synthetic data on JGEX-AG-231~\citep{alphageometry} and IMO-AG-30~\citep{alphageometry}, which consistently outperforms proprietary LLM baselines and is competitive with state-of-the-art systems~\citep{alphageometry} while using orders of magnitude less compute and training data.

\section{Related Work}
\smallsec{Symbolic Approaches}
Classical formal geometry solvers follow two main paradigms~\citep{survey}: synthetic deduction and algebraic computation. Synthetic methods, such as the deductive database approach~\citep{dd,jgex}, employ forward chaining to systematically apply geometric rules and derive new facts, but they struggle with problems requiring complex algebraic manipulation. Algebraic methods, including Gröbner basis~\citep{grobner} and Wu's method~\citep{wu}, encode geometric relations as polynomial equations and solve them algebraically, offering strong reasoning power but often producing proofs that are difficult to interpret. Hybrid systems, such as NGS~\citep{geoqa} and Inter-GPS~\citep{inter-gps} for calculation, LeanEuclid~\citep{leaneuclid} and DD+AR~\citep{alphageometry} for theorem proving, attempt to combine deductive and algebraic reasoning, yet remain specialized in certain task types with limited generality. Other frameworks, including FormalGeo~\citep{formalgeo} and PyEuclid~\citep{pyeuclid}, pursue a unified approach, but still struggle to scale efficiently to Olympiad-level problems.

\smallsec{Datasets and Benchmarks}
Many geometry datasets are derived from textbooks, exercises, and competitions, where problems are paired with manually constructed symbolic formulations~\citep{geoqa_plus,unigeo,pgps9k,formalgeo}. Examples include calculation-oriented datasets such as GeoQA~\citep{geoqa} and Geometry3K~\citep{inter-gps}, as well as theorem proving datasets such as UniGeo~\citep{unigeo}, JGEX-AG-231~\citep{alphageometry}, and IMO-AG-30~\citep{alphageometry}. Due to the high cost of manual annotation, these resources remain relatively small in scale. To expand coverage, recent work~\citep{g-llava,mavis} leverages LLMs and VLMs to generate larger datasets by augmenting existing problems, though their diversity remains bounded by the underlying sources. In parallel, several synthetic pipelines~\citep{geomverse,r-cot,geogen,trustgeogen,geoilp,nesygeo} generate symbolic problems using basic geometric primitives and predicates, but the resulting instances are always constrained in both difficulty and variety. Other benchmarks~\citep{geoeval,mathvision,wemath,geosense}, including MathVista~\citep{mathvista} and MathVerse~\citep{mathverse}, collect a wide variety of geometry questions in natural language to evaluate the reasoning abilities of VLMs. Beyond solving geometry problems, auxiliary datasets have also been introduced for related tasks such as autoformalization~\citep{leaneuclid}, diagram parsing~\citep{pgdp5k}, diagram understanding~\citep{autogeo}, and geometric image generation~\citep{geogpt4v}.

\smallsec{Learning-Based Methods}
Recent advances in LLMs and VLMs have spurred a wave of learning-based approaches to geometric reasoning~\citep{survey1,survey2}. One line of work adopts neuro-symbolic methods that operate over symbolic representations~\citep{geoqa,inter-gps,unigeo,geodrl,laji,g-llava,e-gps,alphageometry,tonggeometry,rep,pi-gps,dfe-gps,autogps}: these approaches leverage LLMs or VLMs to generate solution steps in symbolic form and delegate execution to a solver, ensuring both correctness and interpretability. For example, Inter-GPS~\citep{inter-gps} predicts program sequences to compute numerical quantities, while AlphaGeometry~\citep{alphageometry} predicts auxiliary constructions and integrates them with its symbolic engine for theorem proving. Such systems enable faithful reasoning but heavily rely on the solver's design and symbolic coverage. In contrast, purely neural methods reason directly in natural language~\citep{g-llava,geo-llava,r-cot,nesygeo}, typically targeting calculation problems with verifiable numerical answers and using chain-of-thought reasoning~\citep{cot} to produce step-by-step solutions. While effective on simple problems, these approaches often generate hallucinated intermediate steps and are difficult to verify, limiting their reliability for theorem proving and other complex scenarios. Some methods~\citep{sca-gps,lans,eagle,geox,geodano} also aim to improve visual perception and diagram understanding in VLMs for geometric reasoning. Nevertheless, no unified framework yet exists for training LLMs and VLMs to flexibly support diverse geometry tasks across both formal and natural languages.

\section{Method}
\label{sec:method}

\begin{figure*}[t]
\centering
\includegraphics[width=0.8\linewidth]{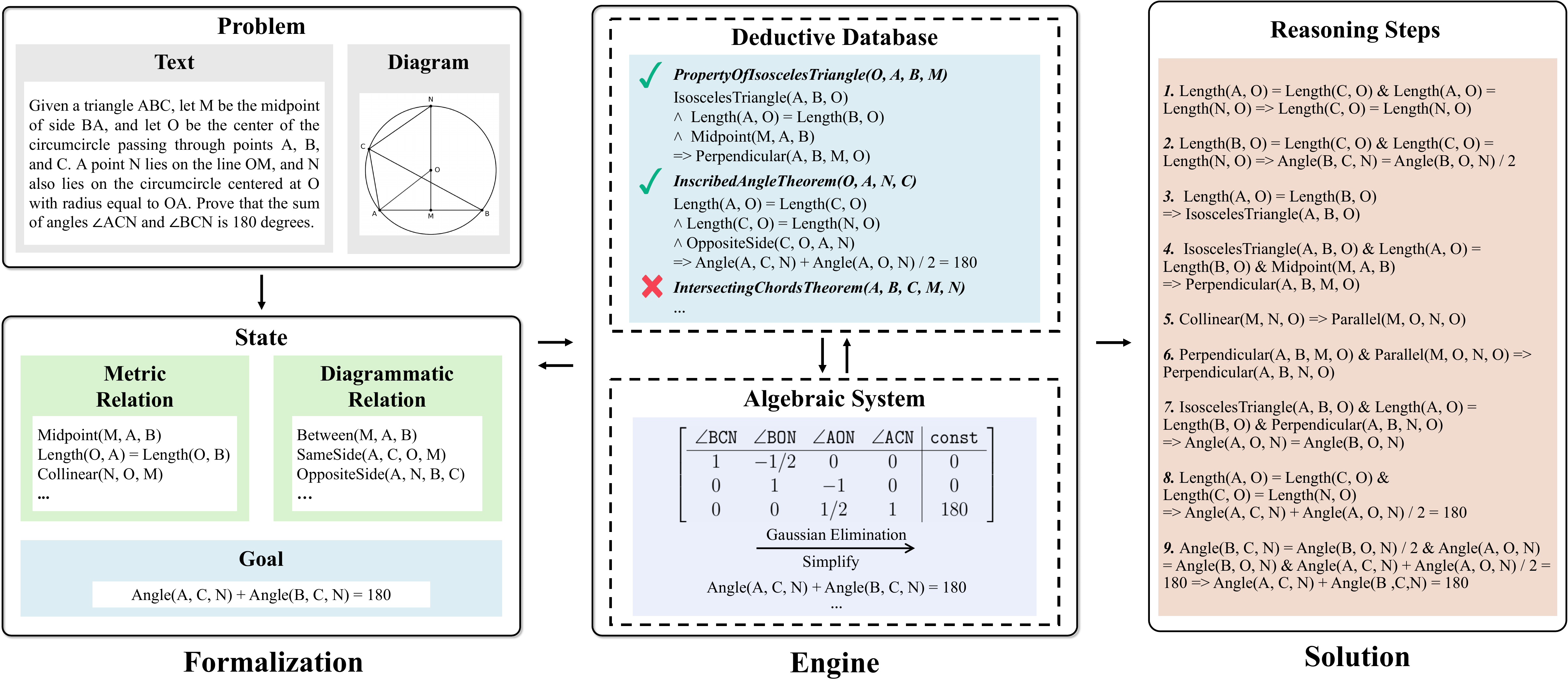}
\vspace{-0.5em}
\caption{An illustrative example of \solver{} solving a proving-style geometry problem.
}
\vspace{-0.75em}
\label{fig:solver}
\end{figure*}

\subsection{Euclidea}
\solver{} is a Python-based formal plane geometry system that encodes information from both text and diagrams. Its reasoning engine integrates deductive inference with algebraic computation. An overview of \solver{} is shown in Figure~\ref{fig:solver}.

\smallsec{Problem Formalization}
\solver{} formalizes plane geometry by combining two established approaches to geometric representation~\citep{dd,e}~\ref{app:formalization}. It treats points as the basic primitives, while all other objects (e.g., lines, triangles) are defined in terms of points~\citep{dd}. A diagram is then formalized as a set of points together with their relations, which fall into two categories~\citep{e}:
\begin{itemize}[nosep, leftmargin=*]
    \item \textit{Metric relations} encode quantitative properties, such as \texttt{Perpendicular(a,b,c,d)} (\texttt{ab}$\perp$\texttt{cd}) or $\texttt{Angle(a,b,c)}=\pi/2$ ($\angle abc = 90^\circ$), and include algebraic equations over geometric quantities such as lengths, angles, ratios, and areas.
    \item \textit{Diagrammatic relations} capture topological configurations that can be read off directly from the diagram, such as \texttt{SameSide(a,b,c,d)} (points \texttt{a} and \texttt{b} lie on the same side of line \texttt{cd}).
\end{itemize}
As in human reasoning, metric relations must be stated explicitly---either given in the problem statement or derived through geometric theorems---whereas diagrammatic relations are typically implicit and inferred from the diagram.

\smallsec{Reasoning Engine}
Given a diagram, \solver{} combines a deductive database with an algebraic system to derive new relations from the initial conditions.
The deductive component extends existing approaches~\citep{jgex,alphageometry} with a richer and more fine-grained set of inference rules defined over our formal representations, and these rules are systematically enumerated to identify applicable theorems. For example, the Angle Bisector Theorem can be formalized as:
\begin{flushleft}
\texttt{AngleBisectorTheorem(a,b,c,d): Angle(d,a,b) = Angle(d,a,c) $\land$ Collinear(d,b,c) $\land$ Between(d,b,c) $\land$ Not(Collinear(a,b,c)) $\Rightarrow$ Length(d,b)/Length(d,c) = Length(a,b)/Length(a,c)}
\end{flushleft}
where point \texttt{d} lies on line \texttt{bc} and on the angle bisector of $\angle bac$.
Note that, unlike previous approaches~\citep{dd} which treat equal-angle as an atomic proposition, \solver{} treats it as a relation between interpreted variables, which is key to bridging geometry and algebra and to handling both calculation and proving tasks.
An SQL database~\citep{sqlite} is used to efficiently enumerate applicable rules: relations and equivalence classes of variables are stored in tables, and conditions are represented as table joins.
This design makes the deductive database easily extensible: new inference rules can be added without requiring manual implementation of dedicated enumerators.
Further details are provided in Appendix~\ref{app:dd}.

Complementing the deductive database, \solver{} integrates a symbolic algebraic system built on SymPy~\citep{sympy} to simplify equations and solve for unknown quantities.
Inspired by DD+AR~\citep{alphageometry}, equations are categorized into four types: (i) \emph{angle-based} (fixed angle, angle sum, and angle ratio), (ii) \emph{length-based} (fixed length, length sum, and length ratio), (iii) \emph{length-ratio-based} (fixed length, length ratio, and equalities between ratios or between an area and the product of two lengths), and (iv) \emph{complex} (all remaining forms, e.g., trigonometric or higher-order polynomial ones).
The first two types can be transformed into a linear system $A\mathbf{x} = \mathbf{b}$, where $\mathbf{x}$ is a vector of geometric quantities and $A$ and $\mathbf{b}$ denote the corresponding coefficients and constants, which is then solved via Gaussian elimination.
The third type can be reduced to a log-linear form and solved analogously.
For typical geometry problems, the diagram retains some degrees of freedom and the known conditions are underdetermined.
Rather than producing a unique solution, Gaussian elimination reduces the system to a minimal set of free variables and expresses all other variables in terms of them.
We can then determine whether an expression evaluates to a constant, or whether an equation holds, by substituting the values into the equation and checking whether the free variables can be eliminated.
For complex equations, \solver{} leverages the accumulated linear/log-linear results for simplification and substitution, often reducing them to single- or double-variable equations that can be further simplified into new relations or solved directly for unknown quantities. Some examples are provided in Appendix~\ref{app:as}.

\solver{} invokes the deductive database and the algebraic system in tandem, with each component reinforcing the other, incrementally expanding the state with new relations. A problem is solved once the goal is either contained in the derived relations or evaluated to a numerical value. The engine halts once closure is reached and no further conclusions can be derived.

\smallsec{Solution Generation}
To generate human-readable reasoning traces, each relation produced by the deductive database is labeled with its originating inference rule and conditions.
For each equation $e$ solved via Gaussian elimination, we cast the tracking process as an optimization problem:
\begin{equation*}
    \min_{\mathbf{z}} \; \|\mathbf{z}\|_{t}, \quad \text{s.t.} \quad [A \; | \; \mathbf{b}]^\top \mathbf{z} = \mathbf{c},
\end{equation*}
where $[A \; | \; \mathbf{b}]$ is the augmented coefficient matrix of the linear system $A\mathbf{x} = \mathbf{b}$, $\mathbf{c}$ is the coefficient vector of the query equation $e$, $\mathbf{z}$ denotes the coefficients of the equations contributing to the query, and $t$ specifies the chosen norm (0 or 1) to promote sparsity and yield a minimal set of traced equations. The optimization is solved with PySCIPOpt~\citep{pyscipopt}.
For quantities derived from complex equations, \solver{} records the original complex equation together with the substituted equations from the previous linear systems.
Detailed examples are provided in Appendix~\ref{app:sg}.

Starting from the goal, \solver{} recursively traces all dependent relations until each is grounded in the initial conditions.
This yields a dependency graph in which the goal is the root, the given conditions are the leaves, the inferred relations form the intermediate nodes, and the edges represent their dependencies.
A post-order traversal then linearizes these relations into an ordered sequence of reasoning steps, which is formatted into a coherent, human-readable symbolic solution.

\begin{figure*}[t]
\begin{center}
\includegraphics[width=0.8\linewidth]{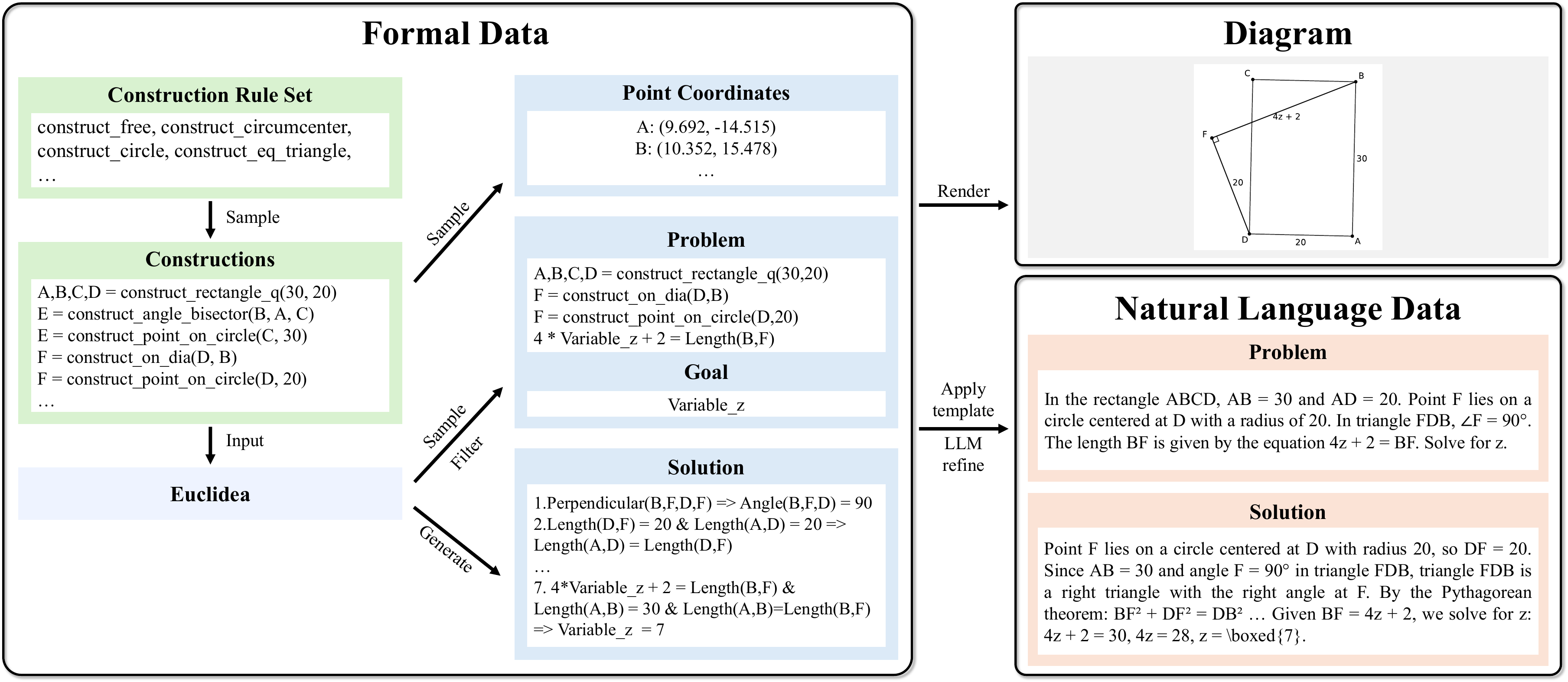}
\end{center}
\vspace{-0.5em}
\caption{An illustrative example of \framework{} generating a calculation-style geometry problem.}
\vspace{-0.75em}
\label{fig:framework}
\end{figure*}

\subsection{Euclid-Omni}
Building on the formalization and reasoning of \solver{}, \framework{} provides a unified framework that integrates a synthetic problem generator, a diagram renderer, and a natural language translator to produce large-scale, diverse training data with flexible configurations for a variety of geometry tasks. An overview of \framework{} is shown in Figure~\ref{fig:framework}.

\smallsec{Synthetic Problem Generation}
\framework{} synthesizes plane geometry problems \emph{from scratch}, allowing fine-grained control over problem structure. Inspired by AlphaGeometry~\citep{alphageometry}, we extend a library of artificial \emph{construction rules}, each of which corresponds to a ruler-and-compass operation that bundles a set of condition and conclusion relations to construct new points, optionally conditioned on previously constructed ones. For example, \texttt{x = construct\_foot(a,b,c)} constructs the foot of the perpendicular from point \texttt{a} to line \texttt{bc}, and is formalized as:
\begin{flushleft}
\texttt{$\exists$ x, Not(Collinear(a,b,c)) $\Rightarrow$ Perpendicular(x,a,b,c) $\land$ Collinear(x,b,c)}
\end{flushleft}

To generate a new problem, \framework{} iteratively applies construction rules until the desired complexity is reached.
For each selected rule, \framework{} samples numerical coordinates that satisfy all required conditions, and then adds the corresponding conclusions to the current state of \solver{}.
To support calculation-style problems, \framework{} also parameterizes a subset of construction rules with explicit geometric quantities such as lengths or angles. For example, \texttt{a,b,c,d = construct\_square\_q(l)} constructs a square \texttt{abcd} with side length \texttt{l}, represented as:
\begin{flushleft}
\texttt{$\exists$ a,b,c,d, True $\Rightarrow$ Square(a,b,c,d) $\land$ Length(a,b) = l}
\end{flushleft}
Notably, even when the same sequence of construction rules is applied, the randomly sampled point coordinates can give rise to distinct topological configurations across runs. An illustrative example is provided in Appendix~\ref{app:construction_rules}.

Given a sampled diagram, \solver{} infers all possible conclusions from the construction and generates the corresponding reasoning steps.
To generate a problem from a diagram, we filter and select specific conclusions as target goals according to the task requirements.
For each selected goal, we trace its minimal supporting set of construction rules from the inferred solution and remove any redundant constructions, thereby preserving the structural minimality of the synthesized problem.

\smallsec{Diagram Rendering}
We implement a diagram renderer that visualizes each sampled problem by drawing geometric objects such as segments and circles from their point coordinates.
For each construction rule, we predefine the set of objects and annotations to display, and render them on a Matplotlib~\citep{matplotlib} canvas.
For example, \texttt{x = construct\_foot(a,b,c)} draws the segments \texttt{bc}, \texttt{xa}, \texttt{xb}, and \texttt{xc}, as well as the right angle \texttt{$\angle$axb}.

\smallsec{Natural Language Translation}
To train models to perform geometric reasoning in natural language, \framework{} supports translating symbolic problems and solutions into fluent textual form.
However, translating directly via prompting would require specifying the full formal language and its semantics, which inevitably leads to lengthy prompts and offers no guarantees of correctness or consistency.
Instead, we adopt a hybrid strategy~\citep{autogeo}: we construct a library of manually verified natural language templates that cover each construction rule appearing in the problem and each relation appearing in the solution.
Given a symbolic problem and its solution, we parse their structure and instantiate the templates to produce aligned drafts, and then prompt an LLM to paraphrase these drafts into more natural and diverse problem statements and reasoning steps.
Examples of templates and prompts are provided in Appendix~\ref{app:informalization}.

\smallsec{Task Configuration}
With \framework{}, we can generate formal and natural language problems together with their corresponding solutions and diagrams.
The pipeline allows users to flexibly configure each component to produce data tailored to specific geometric reasoning tasks.
In this paper, we focus on two representative settings:
(i) following standard visual reasoning benchmarks~\citep{mathvista,mathverse}, we evaluate VLMs on calculation problems solved end-to-end from natural-language statements and diagrams; and
(ii) following IMO-oriented systems~\citep{alphageometry,alphageometry2}, we adopt a neuro-symbolic workflow in which an LLM proposes auxiliary constructions in a formal language and a symbolic solver integrates these candidates to solve Olympiad-level proving problems.

For the first setting, the problem generator samples construction rules, optionally with quantitative parameterization, and restricts target goals to those involving geometric quantities such as lengths, angles, and areas.
Variable-based formulations are also supported by defining linear equations over lengths or angles (e.g., $x + 10^\circ = \angle abc$) and treating the variable as the goal.
To support the multiple-choice format used by existing benchmarks, we adapt the LLM prompts during natural language translation to generate plausible distractors of a scale comparable to the correct answer.
For training, we translate the generated problem and solution into natural language using the methods described above.
The supervised fine-tuning template is:
\texttt{Inputs: <diagram> <natural language problem> \quad Outputs: <natural language solution> \textbackslash boxed\{<final answer/choice>\}}.

For the theorem proving setting, auxiliary constructions are constructions that are necessary for proving the target goal but are not directly involved in constructing it.
Finding such auxiliary constructions is typically the most challenging part of competition-level theorem proving.
Therefore, when training LLMs for theorem proving, we remove the auxiliary constructions from the problem and train the model to recover them.
We sample 8--10 construction rules and restrict goals to common Olympiad-style targets such as midpoint, collinearity, similarity, congruence, concyclicity, and equality of lengths or angles.
Following prior work~\citep{tonggeometry}, only the formal problem statement and the auxiliary constructions are retained for training.
The supervised fine-tuning template is:
\texttt{Inputs: <formal problem> \quad Outputs: <formal auxiliary constructions>}.

We provide several examples of synthetic instances for these two tasks in Appendix~\ref{app:synthetic_examples}. Note that the \framework{} pipeline can also be configured to generate data for other tasks, such as autoformalization~\citep{leaneuclid}, diagram generation~\citep{geouni}, and diagram understanding~\citep{autogeo}. We discuss these potential applications in Section~\ref{sec:discussion} and leave them as future directions for the community to explore.

\section{Experiments}
\label{sec:experiments}
\subsection{Symbolic Solvers}
\smallsec{Setup}
We evaluate \solver{} against open-source formal geometry systems on three benchmarks: (i) Geometry3K~\citep{inter-gps}, SAT-style calculation problems; (ii) JGEX-AG-231~\citep{alphageometry}, textbook theorems and Olympiad-level problems; and (iii) IMO-AG-30~\citep{alphageometry}, IMO problems from 2000--2022. For Geometry3K, we adopt the PyEuclid formalization~\citep{pyeuclid} with minor modifications and correct several annotation errors in the released logical forms. For JGEX-AG-231 and IMO-AG-30, we adopt the original formalizations with minor adjustments for compatibility with our formal language.
For calculation tasks, we compare against Inter-GPS~\citep{inter-gps} and PyEuclid~\citep{pyeuclid}; for proving, against DD+AR~\citep{alphageometry}, Newclid~\citep{newclid}, and PyEuclid~\citep{pyeuclid}. Following PyEuclid's protocol, a numerical prediction is correct if within 2\% of the labeled answer, while a proving solution must output a valid proof. We use a 600-second time limit per problem.

\begin{table}[t]
    \centering
    \caption{Solved problems by formal geometry solvers on three calculation and proving benchmarks (totals in parentheses). --: unsupported task or formalization; $^\dagger$: reported in prior work.}
    \label{tab:solvers}
    \small
    \setlength{\tabcolsep}{4pt}
    \begin{tabular}{lccc}
    \toprule
    \multirow{2}{*}{Solver}
    & \multicolumn{1}{c}{Calculation}
    & \multicolumn{2}{c}{Proving} \\
    \cmidrule(lr){2-2}\cmidrule(lr){3-4}
    & Geometry3K (601) & JGEX-AG-231 (231) & IMO-AG-30 (30) \\
    \midrule
    Inter-GPS$^\dagger$ & 426 & --  & -- \\
    PyEuclid$^\dagger$  & 567 & 202 & -- \\
    DD+AR$^\dagger$     & --  & 198 & 14 \\
    Newclid             & --  & 188 & 14 \\
    \midrule
    \solver{}           & \textbf{595} & \textbf{207} & \textbf{16} \\
    \quad w/o algebraic system     & 1   & 74  & 0 \\
    \quad w/o deductive database   & 36  & 2   & 0 \\
    \bottomrule
    \end{tabular}
    \vspace{-0.75em}
\end{table}

\smallsec{Results}
Table~\ref{tab:solvers} shows that \solver{} consistently outperforms all existing systems, solving 99\% of Geometry3K and two additional challenging IMO problems. The ablation confirms that the deductive database and the algebraic system are individually insufficient; their integration is crucial and enables \solver{} to solve an order of magnitude more problems. The remaining unsolved problems fall into three categories: those that cannot be formalized within \solver{}, outliers with too many points (yielding prohibitively large search spaces and timeouts), and those requiring auxiliary constructions. Beyond higher solve rates, \solver{} also produces higher-quality symbolic proofs that are more human-like and better aligned with diagrams than those of DD+AR and Newclid, while remaining significantly more compact than those of PyEuclid (Appendix~\ref{app:proof}).

\subsection{Natural-Language Calculation Problems}
\smallsec{Setup}
Using \framework{}, we synthesize 10K training instances and translate each into natural language with Gemini~2.5~Flash~\citep{gemini}, which serves as the LLM component of \framework{}. Since benchmark diagrams may include non-geometric objects such as buildings, trees, or tables, we follow prior work~\citep{geogen,r-cot} and augment with 10K examples randomly sampled from Geo170K~\citep{g-llava} to better match the benchmark distribution. We fine-tune Qwen2.5-VL~\citep{qwen2.5vl} on the combined 20K examples for 3 epochs on 8$\times$H100 GPUs using LLaMA-Factory~\citep{llamafactory}.
We compare against G-LLaVA~\citep{g-llava}, MAVIS~\citep{mavis}, Qwen2.5-VL~\citep{qwen2.5vl}, and recent Qwen2.5-VL-based methods (GeoGen~\citep{geogen}, TR-COT~\citep{r-cot}, NeSyGeo~\citep{nesygeo}), all of which are trained via supervised fine-tuning on synthesized data. We evaluate on four benchmarks: GeoQA~\citep{geoqa}, Geometry3K~\citep{inter-gps}, and the plane-geometry subsets of MathVista~(testmini)~\citep{mathvista} and MathVerse~(vision-intensive)~\citep{mathverse}, reporting multiple-choice accuracy when options are provided and exact match to the ground-truth value otherwise.

\begin{table}[t]
    \centering
    \caption{Accuracy (\%) of VLMs on four calculation benchmarks. --: not reported in prior work.}
    \label{tab:task1}
    \small
    \setlength{\tabcolsep}{4pt}
    \begin{tabular}{lccccc}
    \toprule
    Model & \#Train & GeoQA & Geometry3K & MathVista & MathVerse \\
    \midrule
    \textit{Prior VLM baselines} \\
    G-LLaVA-7B & 117K & 64.2 & --   & 53.4 & -- \\
    MAVIS-7B   & 834K & --   & --   & 64.1 & 27.9 \\
    \midrule
    \textit{Qwen2.5-VL variants} \\
    Qwen2.5-VL-7B & --   & 69.4 & 56.4 & 72.2 & 44.1 \\
    + NeSyGeo     & 100K & 71.8 & --   & --   & 46.7 \\
    + GeoGen      & 224K & \textbf{77.6} & 58.4 & 74.0 & -- \\
    + TR-COT      & 183K & \textbf{79.2} & --   & 74.5 & -- \\
    \midrule
    + Ours & 20K  & 76.6 & \textbf{61.0} & \textbf{74.7} & \textbf{51.0} \\
    \bottomrule
    \end{tabular}
    \vspace{-0.75em}
\end{table}

\smallsec{Results}
Table~\ref{tab:task1} shows that our model achieves state-of-the-art accuracy on three of the four datasets, improves over the base model by 5.3\% on average, and remains competitive on GeoQA---all with orders of magnitude less training data than prior methods. Since our rendered diagrams differ substantially from those in the evaluation benchmarks, this also underscores the generalization enabled by our synthetic data. Section~\ref{sec:synthetic_data} analyzes the synthetic dataset, Appendix~\ref{app:mixed_datasets} ablates the training mix, and Appendix~\ref{app:informal_calc} provides qualitative comparisons against the base model.

\subsection{Symbolic Olympiad-level Proving Problems}
\smallsec{Setup}
We synthesize 100K training problems that require auxiliary constructions and fine-tune Qwen2.5-Math-7B~\citep{qwen2.5math} for one epoch on 8$\times$H100 GPUs with LLaMA-Factory~\citep{llamafactory}. At inference time, we use beam search to propose auxiliary constructions: candidates are ranked by log probability, and the search is progressively expanded whenever the solver fails to reach the target goal. We set the branching factor to 32, the beam size to 128, and the maximum depth to 4.
As baselines, we include AlphaGeometry~\citep{alphageometry} and two prompted proprietary models, GPT-4o and Gemini~2.5~Flash, chosen for strong reasoning performance and favorable cost under the high query volume induced by beam search; both are prompted with the formal semantics of DD+AR~\citep{alphageometry} and \solver{} to generate auxiliary constructions. We evaluate on JGEX-AG-231~\citep{alphageometry} and IMO-AG-30~\citep{alphageometry} with a 90-minute timeout per problem, matching the standard IMO duration.

\begin{table}[t]
\centering
\caption{Solved problems by neuro-symbolic systems on two proving benchmarks. --: model/result is not released/reported.}
\label{tab:task3}
\small
\setlength{\tabcolsep}{4pt}
\begin{tabular}{lllcc}
\toprule
Engine & Model & \#Train & JGEX-AG-231 & IMO-AG-30 \\
\midrule
\multirow{4}{*}{DD+AR}
& GPT-4o              & --    & 213 & 17 \\
& Gemini~2.5~Flash     & --    & 216 & 17 \\
& AlphaGeometry        & 100M  & 228 & 25 \\
& AlphaGeometry        & 20M   & --  & 21 \\
\midrule
\multirow{3}{*}{\solver{}}
& GPT-4o              & --    & 213 & 17 \\
& Gemini~2.5~Flash     & --    & 213 & 17 \\
& Ours & 100K  & \textbf{223} & \textbf{22} \\
\bottomrule
\end{tabular}
\vspace{-0.75em}
\end{table}

\smallsec{Results}
Prompting proprietary models yields only modest gains and typically solves easier instances that require at most one auxiliary construction. In contrast, our hybrid system solves 223 problems on JGEX-AG-231 and 22 on IMO-AG-30---competitive with AlphaGeometry trained on 100M problems and exceeding the 20M variant, despite using only 100K samples, highlighting both the computational efficiency and the data effectiveness of our approach. Beyond aggregate counts, our solver frequently discovers auxiliary constructions that differ from AlphaGeometry's and often reaches a proof in fewer reasoning steps (Appendix~\ref{app:auxiliary_constructions}).

\subsection{Synthetic Data}
\label{sec:synthetic_data}

\smallsec{Setup}
To assess the quality of the data generated by \framework{}, we compare our natural-language calculation dataset against Geo170K~\citep{g-llava}, GeoGen~\citep{geogen}, and TR-CoT~\citep{r-cot} along three axes. For \emph{correctness}, we sample 1K problems and prompt Gemini~2.5~Flash to solve them from the natural-language statements and diagrams, reporting answer accuracy. For \emph{difficulty}, we use solution token length as a proxy for the length of multi-step reasoning chains. For \emph{diversity}, we embed question texts and visualize their distribution with t-SNE on 5K random examples from each dataset.

\smallsec{Results}
\begin{table}[t]
\centering
\caption{Solution length statistics across four synthetic datasets.}
\label{tab:sol_length_stats}
\small
\begin{tabular}{lcccc}
\toprule
 & Geo170K & GeoGen & TR-CoT & Ours \\
\midrule
Average & 109.3 & 210.3 & 86.9 & \textbf{247.5} \\
Median  & 103.0 & 177.0 & 78.0 & \textbf{230.0} \\
\bottomrule
\end{tabular}
\vspace{-0.75em}
\end{table}
Gemini~2.5~Flash solves only 69.6\% of the 1K sampled problems, suggesting that purely VLM-generated solutions contain a substantial fraction of errors. In contrast, when \solver{} provides the symbolic solution and Gemini~2.5~Flash is used only for translation, all instances remain consistent with the ground truth, underscoring the importance of \solver{} for correctness in the \framework{} pipeline.

Table~\ref{tab:sol_length_stats} shows that our dataset has the largest average and median solution lengths, indicating more challenging instances that require longer, multi-step reasoning.


\begin{figure}[t]
\centering
\begin{subfigure}[t]{0.33\linewidth}
  \centering
  \includegraphics[width=\linewidth]{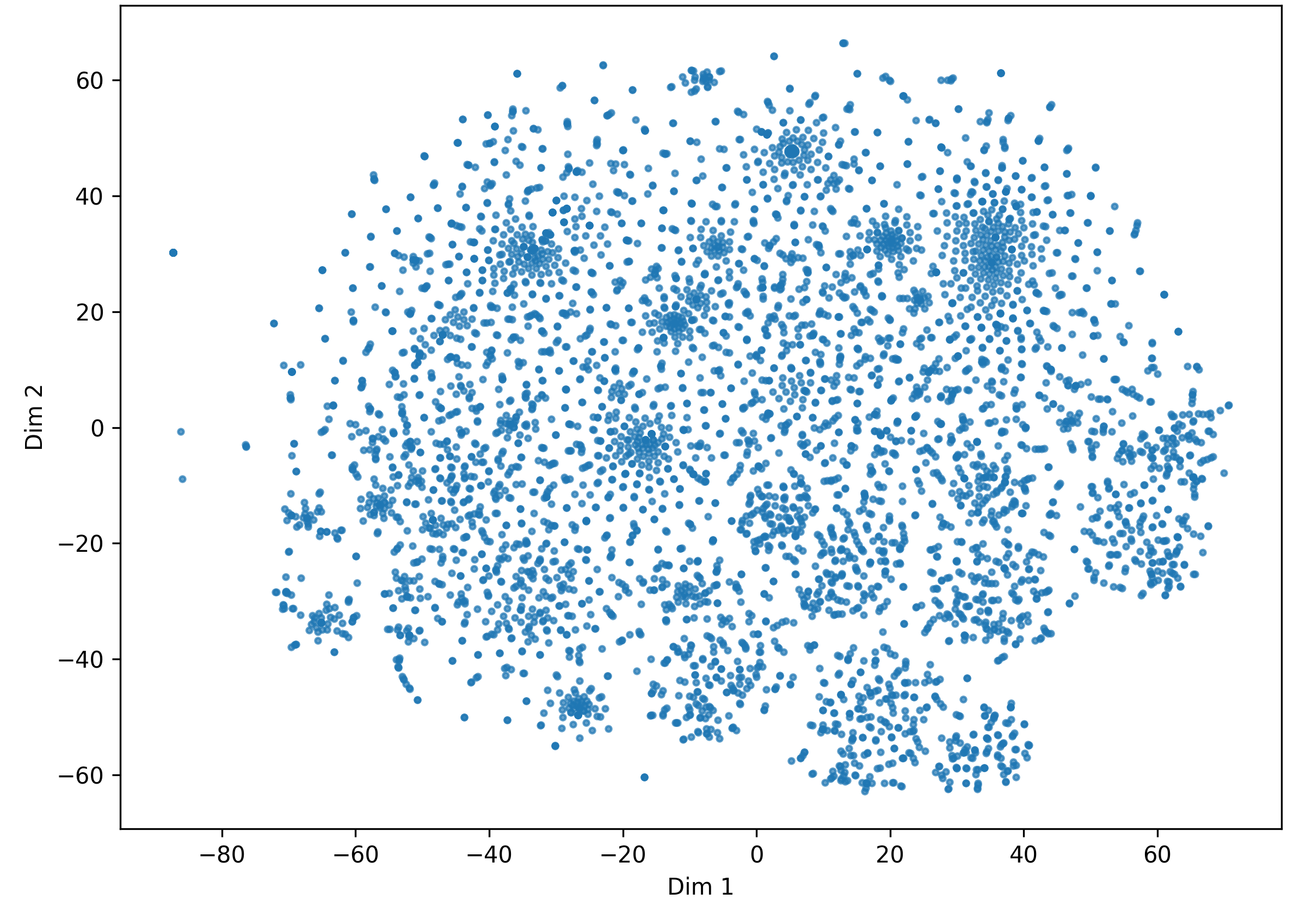}
  \caption{GeoGen}
  \label{fig:tsne_geogen}
\end{subfigure}\hfill
\begin{subfigure}[t]{0.33\linewidth}
  \centering
  \includegraphics[width=\linewidth]{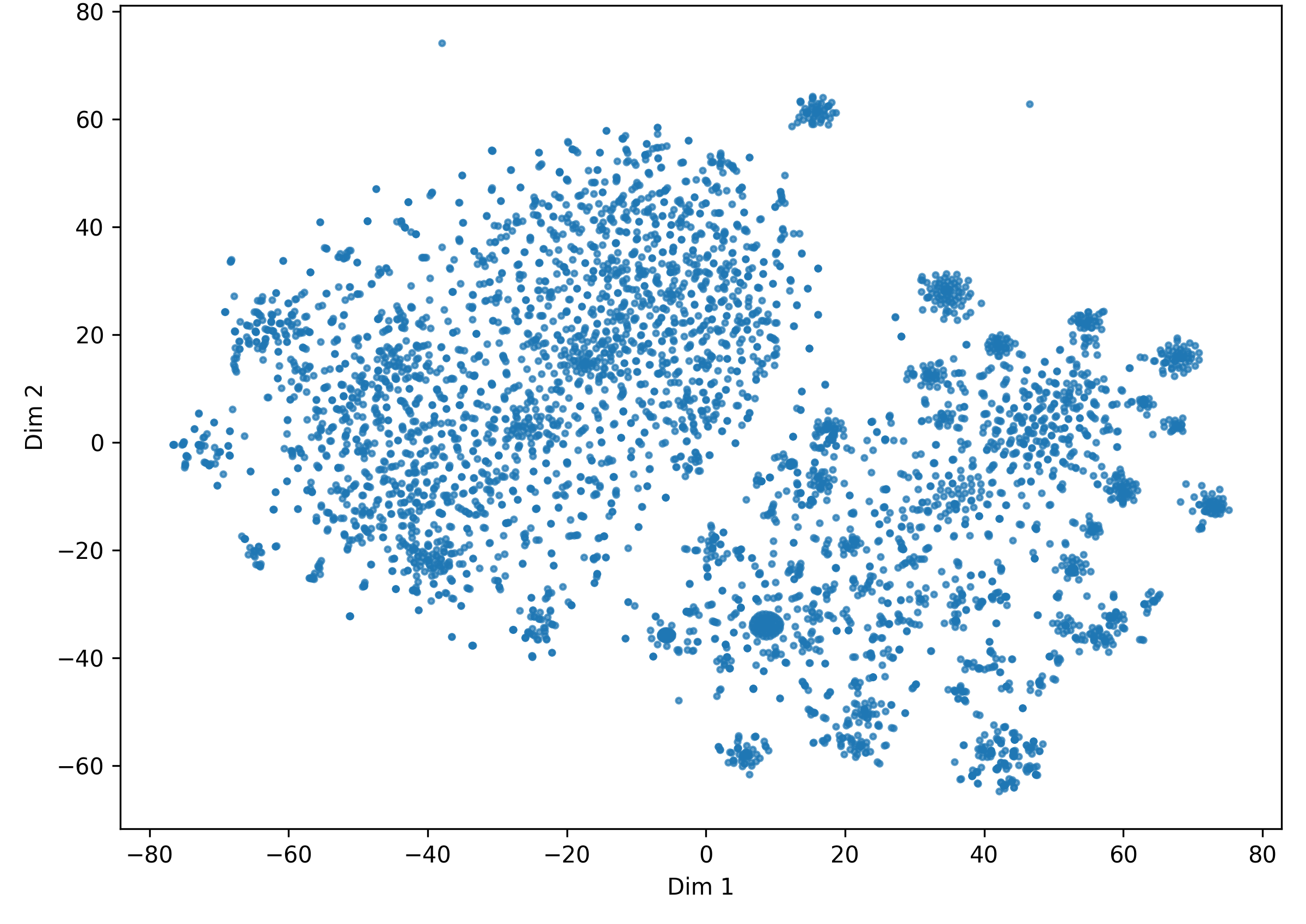}
  \caption{TR-CoT}
  \label{fig:tsne_rcot}
\end{subfigure}\hfill
\begin{subfigure}[t]{0.33\linewidth}
  \centering
  \includegraphics[width=\linewidth]{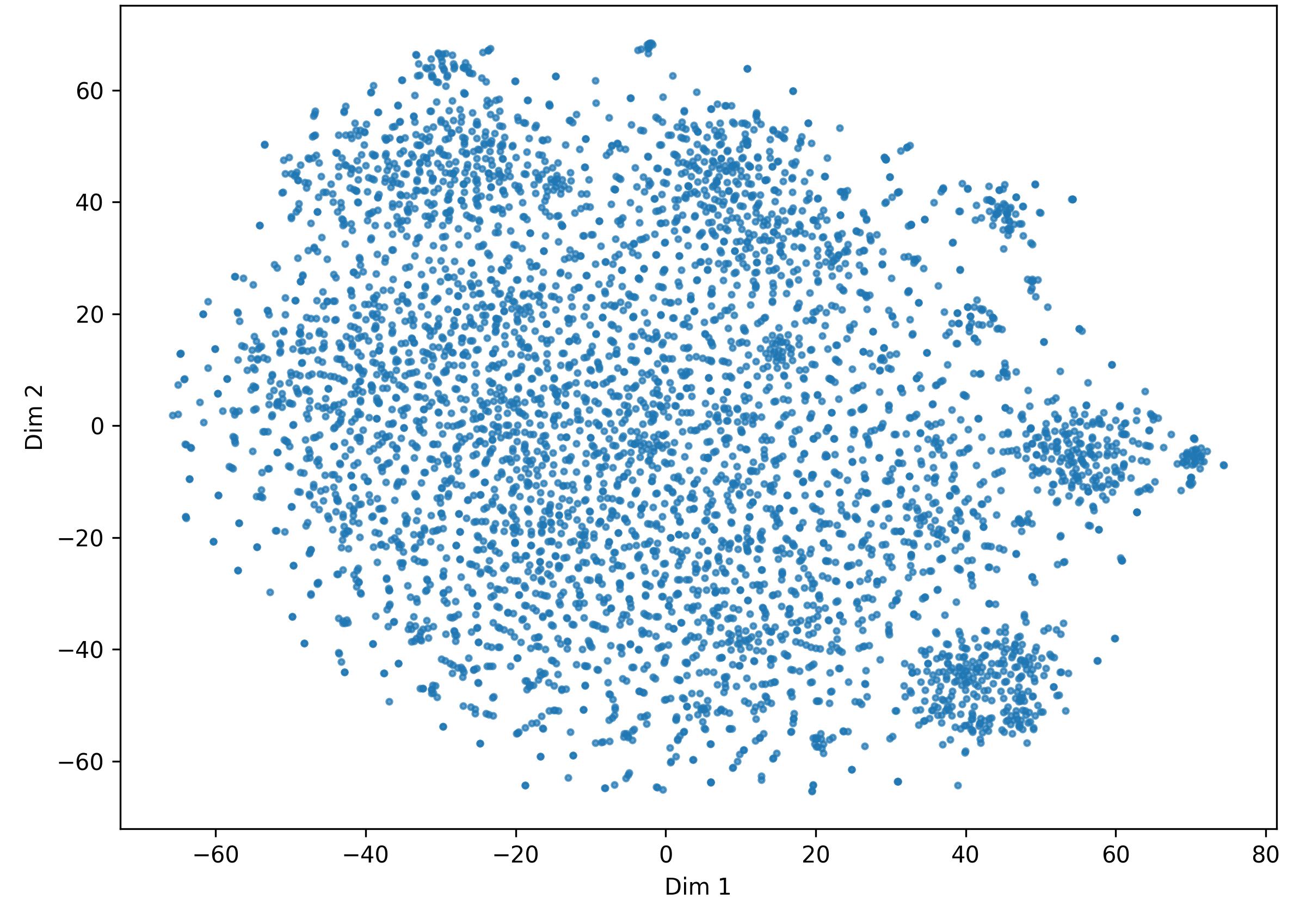}
  \caption{Ours}
  \label{fig:tsne_ours}
\end{subfigure}

\vspace{-0.5em}
\caption{t-SNE visualization of question-text embeddings across three synthetic datasets.}
\vspace{-0.75em}
\label{fig:tsne}
\end{figure}

Figure~\ref{fig:tsne} further shows that our data is more uniformly distributed in the text-embedding space, while other datasets exhibit denser clusters of similar or repetitive samples. Qualitatively, \framework{} also covers a broader set of goal types---angles, lengths, areas, and variable-bound quantities---mirroring real benchmarks, whereas prior pipelines rarely generate area-based or variable-binding goals.


\section{Limitations and Future Work}
\label{sec:discussion}
\smallsec{More Expressive and Human-Like \solver{}}
Although the current design of \solver{} aims to mimic human reasoning and produce human-interpretable solutions, the resulting proofs can be lengthy, particularly for Olympiad-level problems, and differ from those written by IMO contestants. A natural extension is to enrich the deductive database with more sophisticated rules (e.g., well-known theorems) such as Menelaus’ or Desargues’ theorem. This could yield more compact proofs and expand the range of solvable problems. Another direction is to enhance the algebraic system by incorporating projective or inversive geometry, thereby increasing expressivity. Finally, since many primitives of LeanEuclid~\citep{leaneuclid} are already included in \solver{}, a compelling avenue is to ground its representations in Lean~\citep{lean4}. This would allow \solver{} to function as an automatic tactic and integrate with existing libraries~\citep{mathlib,leangeo}, bridging automated geometry solvers with general-purpose proof assistants.

\smallsec{Better Design and Broader Usage of \framework{}}
Currently, construction rules in \framework{} are sampled uniformly, without leveraging human priors or empirical distributions observed in existing problems. Future work could incorporate statistics from existing problems~\citep{tonggeometry} or exploit structural priors such as symmetries to better align generated problems with specific target tasks. In addition, while current \framework{} trains LLMs and VLMs via supervised finetuning, there is significant potential to explore reinforcement learning, using \solver{} to provide verifiable rewards. Beyond the tasks studied here, \framework{} has broader potential applications: it could support autoformalization and problem solving to produce verifiable solutions from natural-language descriptions (useful for pedagogy) or serve as a pretraining resource for VLMs to improve vision–language alignment and diagram understanding.

\section*{Broader Impact}
\label{sec:impact}
This paper presents work whose goal is to advance machine learning for geometric reasoning. We expect its broader impact to be positive, and we do not anticipate societal risks beyond common concerns such as misuse or overreliance.

\bibliography{reference}
\bibliographystyle{abbrvnat}

\newpage
\appendix
\onecolumn
\clearpage

\section{Euclidea}

\subsection{Problem Formalization}
\label{app:formalization}
Following the established framework of geometric formalization introduced by E~\citep{e}, all relations in \solver{} fall into two categories: \emph{metric relations} and \emph{diagrammatic relations}.
A metric relation is encoded either as a predicate proposition (e.g., \(\texttt{Parallel(x,a,b,c)}\)) or as an equation (e.g., \(\texttt{Angle(a,x,b)$= \pi/3$}\)).
A diagrammatic relation asserts certain topological configurations of the diagram, which are encoded as predicate propositions, such as \(\texttt{Between(a,b,c)}\) and \(\texttt{SameSide(a,b,c,d)}\).
These relations are directly extracted from the diagram and are typically not used as target goals in geometry problems. Moreover, for certain relations such as \texttt{Collinear} and \texttt{Between}, their negated forms can also be represented using the \texttt{Not} predicate to assert that the corresponding property does not hold.

Examples of metric relations include:
\begin{flushleft}
\texttt{Collinear(a,b,c)}: points \texttt{a}, \texttt{b}, and \texttt{c} are collinear.\\
\texttt{Parallel(a,b,c,d)}: lines \texttt{ab} and \texttt{cd} are parallel.\\
\texttt{Midpoint(a,b,c)}: point \texttt{a} is the midpoint of segment \texttt{bc}.\\
\texttt{Perpendicular(a,b,c,d)}: lines \texttt{ab} and \texttt{cd} are perpendicular.\\
\texttt{Congruent3(a,b,c,d,e,f)}: triangles \texttt{abc} and \texttt{def} are congruent.\\
\texttt{Similar3(a,b,c,d,e,f)}: triangles \texttt{abc} and \texttt{def} are similar.
\end{flushleft}

Examples of diagrammatic relations include:
\begin{flushleft}
\texttt{Between(a,b,c)}: point \texttt{a} lies between points \texttt{b} and \texttt{c}.\\
\texttt{SameSide(a,b,c,d)}: points \texttt{a} and \texttt{b} lie on the same side of line \texttt{cd}.\\
\texttt{Not(Collinear(a,b,c))}: points \texttt{a}, \texttt{b}, and \texttt{c} are not collinear.\\
\texttt{OppositeSide(a,b,c,d)}: points \texttt{a} and \texttt{b} lie on opposite sides of line \texttt{cd}.
\end{flushleft}

Note that \texttt{Collinear} belongs to metric relations, whereas its negated counterpart \texttt{Not(Collinear)} falls under diagrammatic relations. This distinction arises because collinearity often needs to be formally proved, whereas non-collinearity can typically be inferred directly from the diagram and thus is not treated as a goal in a geometry problem.

It is also worth noting that a key feature of the formalization in \solver{} is its extensibility.  
For \textit{diagrammatic relations}, new predicates can be easily introduced to capture richer geometric semantics. For instance, to indicate whether an angle is acute or obtuse, or whether one segment is longer than another, as inferred directly from the diagram.  
Similarly, the set of \textit{metric relations} is extensible and can incorporate higher-level or composite semantics derived from existing relations by specifying their correspondence with previously defined ones.  
For example,
\[
\texttt{Square(a,b,c,d)} := \texttt{Rectangle(a,b,c,d)} \wedge \texttt{Length(a,b) = Length(a,d)}
\]
Such high-level relations can make the problem formalization more concise and streamline the reasoning steps based on higher-level geometric properties.

Compared with geometric formalization in existing IMO-level systems~\citep{alphageometry,alphageometry2,chen2025seed}, the key innovation of \solver{} lies in its integration of \emph{diagrammatic relations} to support human-like reasoning about angular and topological relationships.

\begin{figure}[htbp]
    \centering
    \begin{subfigure}[t]{0.3\textwidth}
        \centering
        \includegraphics[width=\textwidth]{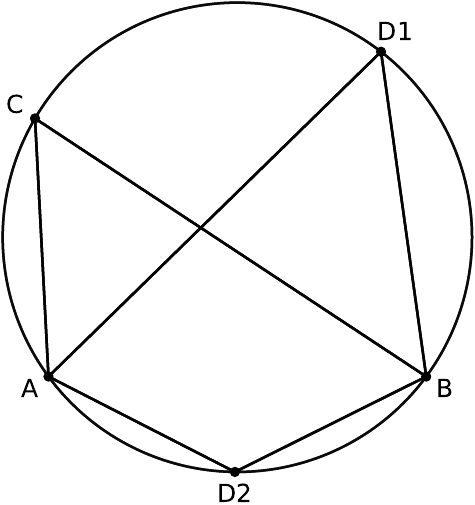}
        \caption{Inscribed angle theorem: 
        \(\texttt{Angle(A,C,B)}\) is equal to \(\texttt{Angle(A,D1,B)}\), 
        but supplementary to \(\texttt{Angle(A,D2,B)}\).}
        \label{fig:inscribed_angle}
    \end{subfigure}
    \hfill
    \begin{subfigure}[t]{0.3\textwidth}
        \centering
        \includegraphics[width=\textwidth]{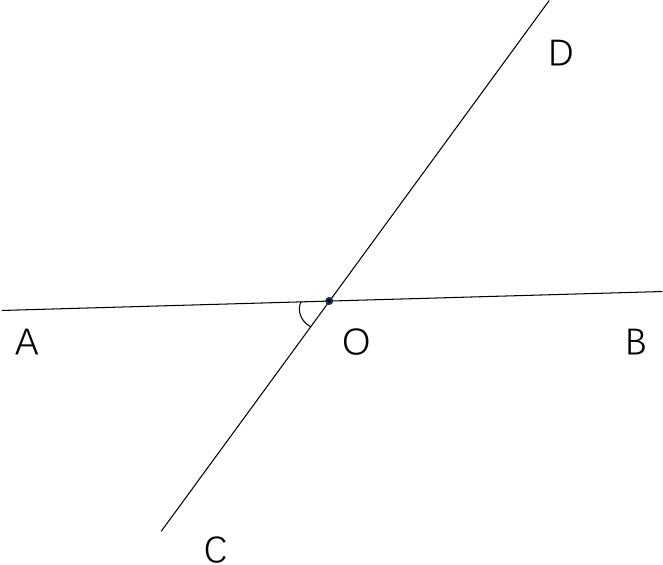}
        \caption{Under the conventional definition of angles:
        \(\texttt{Angle(A,O,C)} = \texttt{Angle(C,O,A)}\), and
        \(\texttt{Angle(A,O,C)}\) is supplementary to \(\texttt{Angle(A,O,D)}\).}
        \label{fig:full_angle}
    \end{subfigure}
    \hfill
    \begin{subfigure}[t]{0.3\textwidth}
        \centering
        \includegraphics[width=\textwidth]{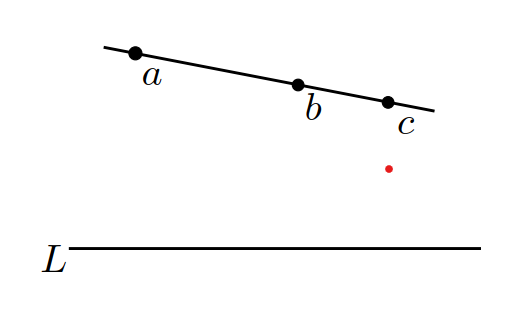}
        \caption{An example of diagrammatic inference from \citet{e}:
        if \(\texttt{b}\) is between \(\texttt{a}\) and \(\texttt{c}\), 
        and \(\texttt{a}\) and \(\texttt{c}\) are on the same side of line \(L\), 
        then \(\texttt{a}\) and \(\texttt{b}\) are on the same side of \(L\).}
        \label{fig:diagrammatic}
    \end{subfigure}
\end{figure}

For humans, two angles are considered equal if and only if the underlying azimuths have the same cosine value.
Consider the inscribed angle theorem illustrated in Figure~\ref{fig:inscribed_angle}.
If points \(\texttt{A,B,C,D}\) lie on the same circle, then \(\texttt{Angle(A,C,B)}\) is either equal or supplementary to \(\texttt{Angle(A,D,B)}\), depending on whether \(\texttt{A}\) and \(\texttt{D}\) lie on the same side or on opposite sides of line \(\texttt{AB}\).

Without diagrammatic relations, however, formal systems such as DD+AR~\citep{alphageometry} cannot distinguish between these two scenarios.
Instead, they adopt the \emph{full-angle notation}~\citep{full-angle}, where two angles are treated as equal if and only if their azimuths have the same sine value.

For instance, in Figure~\ref{fig:full_angle}, under the conventional definition:
\[
\texttt{Angle(A,O,C)} = \texttt{Angle(C,O,A)}, \qquad
\texttt{Angle(A,O,C)} + \texttt{Angle(A,O,D)} = \pi
\]
However, in full-angle notation:
\[
\texttt{Angle(A,O,C)} = \texttt{Angle(A,O,D)}, \qquad
\texttt{Angle(A,O,C)} = -\texttt{Angle(C,O,A)}
\]
This misalignment prevents faithful translation between natural language and formal language.
In theorem-proving settings, the equality of two angles must therefore be expressed as ``equal or supplementary,'' rather than strictly equal.
The issue becomes even more pronounced in calculation-oriented tasks, where such ambiguity prevents the unique determination of an angle’s value and can lead to inconsistent numerical results.
Consequently, it also hinders the formal system’s ability to perform complex algebraic computations, which inherently rely on precise angle representations.

Moreover, diagrammatic inferences, as illustrated in Figure~\ref{fig:diagrammatic}, are rarely made explicit in human geometric reasoning, except occasionally in formal logic.
Therefore, we extract them directly from the diagram as part of the initial conditions.
Integrating these inference rules into the deduction system would increase its complexity by introducing reasoning through contradiction and disjunction, and would also make proofs unnecessarily verbose and less aligned with human reasoning patterns.

In contrast to the formal system E~\citep{e}, which distinguishes three types of geometric objects—points, lines, and circles—\solver{} treats \emph{points} as the only first-class entities, assuming all points are implicitly connected.
For example, our assertion
\[
\texttt{Collinear(a,b,c)}
\]
is equivalent to the formal formulation in E~\citep{e}:
\[
\texttt{a,b,c: Point}, \quad \texttt{l: Line}, \quad \texttt{on(a,l)}, \; \texttt{on(b,l)}, \; \texttt{on(c,l)}
\]

This simplification reduces the complexity of our rule-based deduction system and avoids trivial auxiliary constructions such as ``connect point \(\texttt{a}\) and point \(\texttt{b}\).''

\subsection{Deductive Database}
\label{app:dd}
Given the current state, \solver{} stores all geometric entities (i.e., points) and predicates (e.g., \texttt{Perpendicular}, \texttt{Collinear}) in an in-memory \texttt{SQL} database, where each predicate is represented as a relational table.
To efficiently manage algebraic dependencies, a \texttt{union-find} data structure is used to maintain equivalence relationships among geometric quantities.
For example, if the system contains \texttt{Length(a,b) = Length(b,c)} and \texttt{Length(b,c) = Length(c,d)}, it can automatically infer \texttt{Length(a,b) = Length(c,d)}.
All such equivalence relations, including those involving angles, angle sums, lengths, and length ratios, are first organized into equivalence classes within \solver{} and then synchronized with the corresponding \texttt{SQL} tables.
Under this design, a conjunction of conditions is naturally expressed as a series of table joins, allowing all applicable inference rules to be enumerated declaratively through \texttt{SQL} queries. 

For example, the condition defining \texttt{Midpoint(a,b,c)} can be expressed as:
\[
\texttt{Length(a,b) = Length(a,c) $\wedge$ Collinear(a,b,c) $\wedge$ Between(a,b,c)},
\]  
which can be translated into the following \texttt{SQL} query:

\begin{tcolorbox}[breakable, enhanced jigsaw]
\small{
\begin{lstlisting}
SELECT a.name AS a, b.name AS b, c.name AS c
FROM points a
JOIN points b
JOIN points c
JOIN length r0l 
  ON ((r0l.p0 = a.name AND r0l.p1 = b.name) OR (r0l.p1 = a.name AND r0l.p0 = b.name))
JOIN length r0r 
  ON ((r0r.p0 = a.name AND r0r.p1 = c.name) OR (r0r.p1 = a.name AND r0r.p0 = c.name))
  AND r0l.component = r0r.component
JOIN collinear r1
  ON ((a.name = r1.p0 AND b.name = r1.p1 AND c.name = r1.p2)
      OR (a.name = r1.p0 AND c.name = r1.p1 AND b.name = r1.p2)
      OR (b.name = r1.p0 AND a.name = r1.p1 AND c.name = r1.p2)
      OR (b.name = r1.p0 AND c.name = r1.p1 AND a.name = r1.p2)
      OR (c.name = r1.p0 AND a.name = r1.p1 AND b.name = r1.p2)
      OR (c.name = r1.p0 AND b.name = r1.p1 AND a.name = r1.p2))
JOIN between r2
  ON ((a.name = r2.p0 AND b.name = r2.p1 AND c.name = r2.p2)
      OR (a.name = r2.p0 AND c.name = r2.p1 AND b.name = r2.p2))
WHERE b.name < c.name;
\end{lstlisting}
}
\end{tcolorbox}

Note that we impose a lexical partial order on variable names to eliminate redundant permutations.
For instance, \texttt{Midpoint(a,b,c)} and \texttt{Midpoint(a,c,b)} describe the same configuration; therefore, the constraint \texttt{b.name < c.name} ensures that only one canonical ordering is retained.

\subsection{Algebraic System}
\label{app:as}
At each iteration, \solver{} first uses Gaussian elimination to solve three types of equations: (i) linear equations of angles, (ii) linear equations of lengths, and (iii) log-linear equations of lengths. The results from these systems are then combined to solve (iv) other nonlinear equations.

Examples of linear equations of angles include:
\begin{flushleft}
\texttt{Angle(d,a,c) = $\pi$/6} \\
\texttt{Angle(a,b,c) + Angle(a,c,b) + Angle(b,a,c) = $\pi$} \\
\texttt{Angle(a,b,c) = 2$\times$Angle(a,c,b)}
\end{flushleft}

Examples of linear equations of lengths include:
\begin{flushleft}
\texttt{Length(a,b) = 3} \\
\texttt{Length(a,m) + Length(b,m) = Length(a,b)} \\
\texttt{Length(a,m) = 2$\times$Length(b,m)}
\end{flushleft}

Examples of log-linear equations of lengths include:
\begin{flushleft}
\texttt{Length(a,b) = 3} \\
\texttt{Length(a,m) = 2$\times$Length(b,m)} \\
\texttt{Length(a,m)/Length(b,m) = Length(a,n)/Length(c,n)} \\
\texttt{Area(a,b,c) = Length(a,b)$\times$Length(a,c)/2}
\end{flushleft}

Examples of non-linear equations include:
\begin{flushleft}
\texttt{Length(a,b)$^2$ + Length(a,c)$^2$ = Length(b,c)$^2$} \\
\texttt{Length(a,b) = Length(b,c)$\times$cos(Angle(a,b,c)) } \\
\texttt{Area(a,b,c,d) = Length(a,b)$\times$Length(b,c)$\times$sin(Angle(a,b,c))}
\end{flushleft}

Although the linear and log-linear equations of lengths involve the same set of variables and could theoretically be solved together, doing so easily leads to high-order polynomials, making symbolic solutions intractable. To avoid this, \solver{} solves the three (log-)linear systems independently and then substitutes their solutions into all remaining non-linear equations, including those involving trigonometric or higher-order polynomials.  

If the resulting equation only involves a single variable, the corresponding geometric quantity has been determined.  
If the equation becomes (log-)linear after substitution, it is reintroduced into the appropriate subsystem. 
This mechanism enables constraint propagation among the three (log-)linear systems without requiring the entire nonlinear system to be solved simultaneously.

\begin{wrapfigure}{r}{0.3\textwidth}
    \centering
    \vspace{-10pt}
    \includegraphics[width=0.9\linewidth]{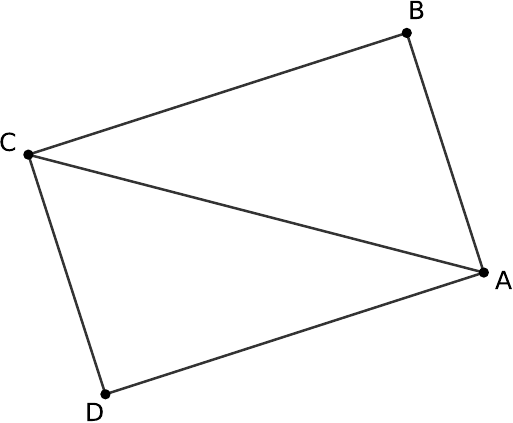}
    \caption{An example diagram of rectangle \texttt{ABCD}.}
    \label{fig:as}
    \vspace{-25pt}
\end{wrapfigure}
As a concrete example, consider Figure~\ref{fig:as}, where the rectangle \texttt{ABCD} is defined with the following conditions:
\[
\texttt{AD + AB} = 7, \quad \texttt{AD - AB} = 1.
\]
Together with the properties of a rectangle, the set of linear equations of lengths can be represented as the following matrix equation:
\[
\begin{pmatrix}
1 & 1 & 0 & 0\\
-1 & 1 & 0 & 0\\
1 & 0 & 0 & -1\\
0 & 1 & -1 & 0
\end{pmatrix}
\begin{pmatrix}
\texttt{AB}\\
\texttt{AD}\\
\texttt{BC}\\
\texttt{CD}
\end{pmatrix}
=
\begin{pmatrix}
7\\
1\\
0\\
0
\end{pmatrix}
\]

First, Gaussian elimination is applied to the system of linear equations to find 
\(\texttt{AB} = 3\) and \(\texttt{BC} = 4\).  
Using the Pythagorean theorem, we have 
\[
\texttt{AC}^2 = \texttt{AB}^2 + \texttt{BC}^2
\]
Substituting the solutions from the linear system, we obtain 
\[
\texttt{AC}^2 = 25
\]
After post-processing the solution and eliminating negative values for lengths, we obtain 
\(\texttt{AC} = 5\). Since the equation \(\texttt{AC} = 5\) satisfies both the linear and log-linear forms of length equations, it is added back to both systems for potential use in subsequent iterations.

\subsection{Solution Generation}
\label{app:sg}
For each relation inferred in the deductive database, \solver{} records its inference rule and conditions as the source.
For instance, consider the SAS (Side–Angle–Side) congruence rule for triangles:
\begin{flushleft}
\texttt{SAS(a,b,c,d,e,f): Not(Collinear(a,b,c)) $\land$ \\
Length(a,b) = Length(d,e) $\land$ Angle(a,b,c) = Angle(d,e,f) $\land$ \\
Length(b,c) = Length(e,f) $\Rightarrow$ Congruent3(a,b,c,d,e,f)}
\end{flushleft}
If the relation \texttt{Congruent3(a,b,c,d,e,f)} is inferred via this rule, its source is the SAS rule, as well as its conditions \texttt{Not(Collinear(a,b,c))}, \texttt{Length(a,b) = Length(d,e)},  
\texttt{Angle(a,b,c) = Angle(d,e,f)}, and \texttt{Length(b,c) = Length(e,f)}.

For the equations inferred from the algebraic system, we adopt different dependency-tracing strategies, since (log-)linear equations are solved using Gaussian elimination, whereas nonlinear equations are solved through substitution and symbolic solving.

Continuing the example in Figure~\ref{fig:as}, suppose \texttt{AC = 5} is the target goal.  
Since it results from substituting \texttt{AB = 3} and \texttt{BC = 4} into \texttt{AC$^2$ = AB$^2$ + BC$^2$},  
the source of \texttt{AC = 5} includes these three equations.  
Next, the system recursively traces the dependencies of \texttt{AB = 3} and \texttt{BC = 4}.

To identify the source of \texttt{AB = 3}, we introduce a coefficient vector 
$\mathbf{z} = (z_1, z_2, z_3, z_4)^\top$, 
where $z_i$ represents the weight assigned to the $i$-th equation that contributes to the query equation \texttt{AB = 3}. 
We then solve the following optimization problem (using the $\ell_0$-norm):
\[
\min_{\mathbf{z}} \; \|\mathbf{z}\|_0 
\quad \text{s.t.} \quad
\begin{pmatrix}
\texttt{AB} & \texttt{AD} & \texttt{BC} & \texttt{CD} & \texttt{const} \\
1 & 1 & 0 & 0 & 7\\
-1 & 1 & 0 & 0 & 1\\
1 & 0 & 0 & -1 & 0\\
0 & 1 & -1 & 0 & 0
\end{pmatrix}^\top
\begin{pmatrix}
z_1 \\ z_2 \\ z_3 \\ z_4
\end{pmatrix}
=
\begin{pmatrix}
1 \\ 0 \\ 0 \\ 0 \\ 3
\end{pmatrix},
\]
where the right-hand vector \((1,\,0,\,0,\,0,\,3)^\top\) encodes the query coefficients corresponding to the equation \(\texttt{AB = 3}\).  
Solving yields \(z_1 = 0.5\), \(z_2 = -0.5\), and \(z_3 = z_4 = 0\),  
indicating that \texttt{AB = 3} depends on the two equations \texttt{AD + AB = 7} and \texttt{AD - AB = 1}.  
A similar process can be applied to identify the source equations for \texttt{BC = 4}. 

Note that the $\ell_0$-norm formulation yields the minimal number of source equations but can be computationally expensive, 
whereas the $\ell_1$-norm provides an efficient approximation that minimizes the total weights but does not always guarantee minimal sparsity.  
In practice, \solver{} can adaptively choose between the $\ell_0$ and $\ell_1$ formulations depending on system complexity and user requirements.

We perform this backtracking process starting from the target goal of the problem and continue until all traced relations and equations are reduced to the initial conditions.
A dependency graph is then constructed, with the goal as the root, the initial conditions as the leaves, and intermediate relations as internal nodes.
A post-order traversal of this graph produces the sequential proof steps.

To make the proofs more human-like and concise, \solver{} merges steps that share common conditions or exhibit hierarchical relationships (e.g., the properties of a square subsume those of a rectangle).
Furthermore, certain diagrammatic relations are omitted from the conditions of proof steps, as they are typically trivial and rarely stated explicitly in human-written proofs.

\section{Euclid-Omni}
\subsection{Problem Synthesis}
\label{app:construction_rules}
To synthesize a geometry problem, \framework{} first samples a sequence of construction rules and then samples a corresponding diagram consistent with these rules.
For calculation problems, parameterized construction rules are used so that the resulting diagram is fully determined by its parameters, up to global translation, rotation, and reflection.
In contrast, for proving problems, the framework also employs construction rules with unspecified degrees of freedom, since the target theorem typically holds across a family of diagrams rather than a single instance.

\begin{figure}[htbp]
    \centering
    \hfill
    \begin{subfigure}[t]{0.35\textwidth}
        \centering
        \includegraphics[width=0.7\textwidth]{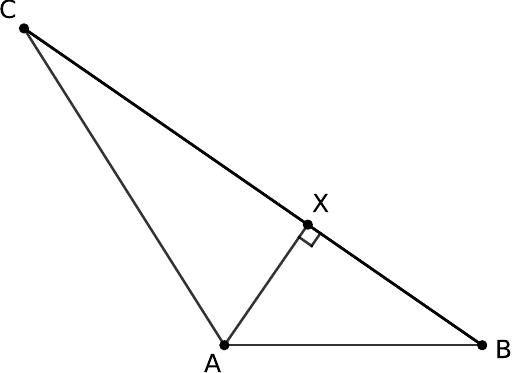}
        \caption{A sampled figure with diagrammatic relations \(\texttt{Between(x,b,c)}\) and \(\texttt{OppositeSide(b,c,a,x)}\).}
        \label{fig:construction_a}
    \end{subfigure}
    \hfill
    \begin{subfigure}[t]{0.35\textwidth}
        \centering
        \includegraphics[width=0.5\textwidth]{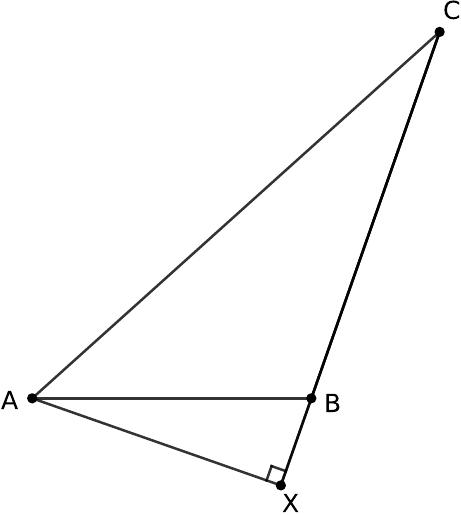}
        \caption{An alternative configuration with \(\texttt{Between(b,x,c)}\) and \(\texttt{SameSide(b,c,a,x)}\).}
        \label{fig:construction_b}
    \end{subfigure}
    \hfill
    \label{fig:mainfigure}
\end{figure}

Consider the constructions illustrated in Figures~\ref{fig:construction_a} and~\ref{fig:construction_b}:
\[
\texttt{a,b,c = construct\_triangle();} \quad \texttt{x = construct\_foot(a,b,c)}
\]
To sample a diagram, the coordinates of all points are drawn according to the specified construction rules. 
Certain conclusions hold for all diagrams generated from these constructions, yielding invariant relations such as 
\(\texttt{Perpendicular(x,a,b,c)}\).
In contrast, diagrammatic relations depend on the specific sampled diagram and can be directly extracted from it. For example, 
\(\texttt{Between(x,b,c)}\) and \(\texttt{OppositeSide(b,c,a,x)}\) in Figure~\ref{fig:construction_a}.
Even with the same construction rules, different diagram samples may yield distinct diagrammatic relations. 
For instance, if \(\texttt{Angle(a,b,c)} > \pi/2\), the resulting diagram instead satisfies 
\(\texttt{Between(b,x,c)}\) and \(\texttt{SameSide(b,c,a,x)}\), as shown in Figure~\ref{fig:construction_b}.

\subsection{Examples of Templates and Prompts for Natural Language Translation}
\label{app:informalization}
Following prior work~\citep{autogeo}, we construct multiple natural language templates for each construction rule (in the problem) and relation (in the solution) within our formalization. When converting a formal expression into natural language, one template is randomly sampled and instantiated to generate the corresponding textual description.
For instance, the construction rule \texttt{x = construct\_circumcenter(a,b,c)} can be verbalized as:
\begin{tcolorbox}[breakable, enhanced jigsaw]
\small{
\begin{lstlisting}[frame=none]
x is the circumcenter of abc
x is the center of the circle passing through a, b, and c
the center of the circle through points a, b, and c is x
the point x is the circumcenter of triangle abc
\end{lstlisting}
}
\end{tcolorbox}
Similarly, each relation can also be expressed through multiple templates.  
For example, the relation \texttt{Perpendicular(a,b,c,d)} can be instantiated as:
\begin{tcolorbox}[breakable, enhanced jigsaw]
\small{
\begin{lstlisting}[frame=none]
line ab is perpendicular to line cd
line ab ⟂ line cd
line through a and b is perpendicular to the line through c and d
the lines formed by (a, b) and (c, d) are perpendicular
\end{lstlisting}
}
\end{tcolorbox}

To refine the problem statement with an LLM, we employ a dedicated prompt that converts the template-based problem and its goal into fluent text. The prompt is defined as follows:
\begin{tcolorbox}[breakable, enhanced jigsaw]
\small{
\begin{lstlisting}[frame=none]
You are given a plane geometry problem:

Problem: <template-based problem>. Determine <template-based goal>.

Task:
- Rewrite the problem in clear, concise, and fluent language, preserving the original meaning.
- Output ONLY the rewritten problem, with no explanations or extra text.
\end{lstlisting}
}
\end{tcolorbox}

Once the refined problem is generated, we employ a separate prompt to rewrite the template-based solution in fluent natural language:
\begin{tcolorbox}[breakable, enhanced jigsaw]
\small{
\begin{lstlisting}[frame=none]
You are given a plane geometry problem and its corresponding solution:

Problem:
<refined problem>

Solution:
<template-based solution>

Task:
- Rewrite the solution in clear, concise, and fluent language, simplifying trivial or redundant steps.
- Step-wise formatting is optional. Use it only when it improves clarity; otherwise, presenting the solution as a continuous paragraph is acceptable.
- Output ONLY the rewritten solution, with the final answer inside \boxed{} at the end.
- Do NOT include the problem statement, explanations, or extra text.
\end{lstlisting}
}
\end{tcolorbox}

We also support converting formal problems and their corresponding solutions into multiple-choice formats, as commonly used in calculation-style geometry datasets~\citep{geoqa,g-llava,mathvista,mathverse}. The prompt for this setting is defined as follows:
\begin{tcolorbox}[breakable, enhanced jigsaw]
\small{
\begin{lstlisting}[frame=none]
You are given a plane geometry problem and its corresponding solution:

Problem: <template-based problem>. <template-based goal> = ( ).

Reference Answer (for correctness only):
<answer>

Task:
- Rewrite the problem in clear, concise, and fluent language, preserving the original meaning.
- Convert the task into a multiple-choice question with exactly 4 options labeled A,B,C,D.
- Use the reference solution ONLY to determine the correct numeric/choice answer.
- Create plausible distractors of comparable scale or magnitude to the correct answer.
- Ensure EXACTLY ONE option is correct.
- Output ONLY the rewritten problem followed by the choices, with no explanations or extra text.
- Do NOT include the solution, rationales, or extra text.

Output format:
<Rewritten problem statement>

Choices:
A: ...
B: ...
C: ...
D: ...
\end{lstlisting}
}
\end{tcolorbox}

Finally, we use a corresponding prompt to refine the solution for the multiple-choice setting:
\begin{tcolorbox}[breakable, enhanced jigsaw]
\small{
\begin{lstlisting}[frame=none]
You are given a plane geometry problem and its corresponding solution:

Problem:
<refined multiple-choice problem>

Solution:
<template-based solution>

Task:
- Rewrite the solution in clear, concise, and fluent language, simplifying trivial or redundant steps.
- Step-wise formatting is optional. Use it only when it improves clarity; otherwise, presenting the solution as a continuous paragraph is acceptable.
- Ensure the final choice label matches the provided solution's final answer.
- Output ONLY the rewritten solution, with the final CHOICE LABEL (e.g., A, B, C, or D) inside \boxed{} at the end.
- Do NOT include the problem statement, explanations, or extra text.
\end{lstlisting}
}
\end{tcolorbox}

\subsection{Examples of Synthetic Data Instances}
\label{app:synthetic_examples}
\begin{figure}[ht]
\begin{center}
\includegraphics[width=0.5\linewidth]{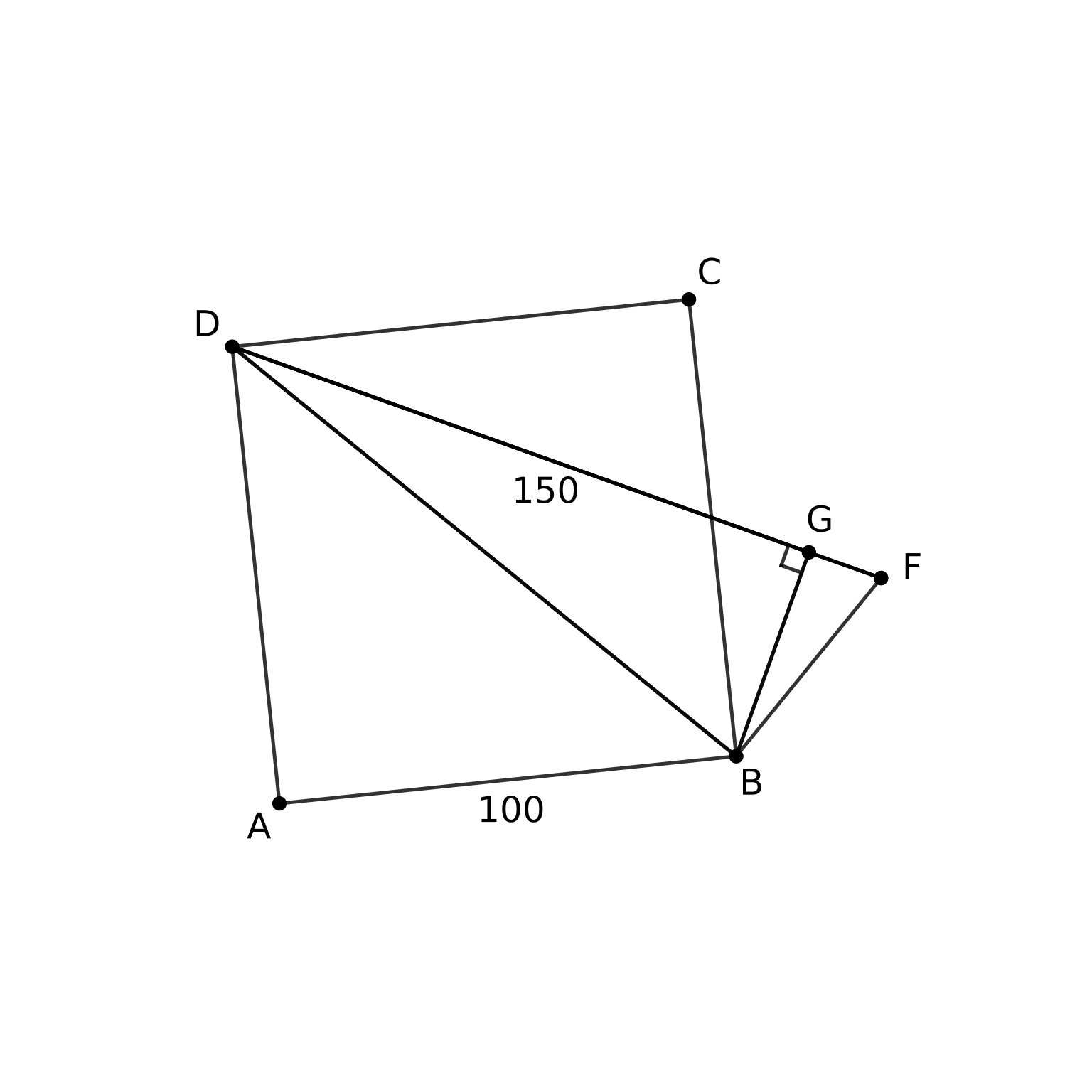}
\end{center}
\vspace{-20pt}
\caption{Diagram of a generated final-answer calculation problem.} 
\label{fig:example1}
\end{figure}

We present two examples of our generated instances for calculation problems: one in a final-answer format and the other in a multiple-choice format. The first example is shown below, with its corresponding diagram provided in Figure~\ref{fig:example1}.
\begin{tcolorbox}[breakable, enhanced jigsaw]
\small{
\begin{lstlisting}[frame=none]
Problem: 
ABCD is a square with side length 100. Point F lies on a circle centered at D with a radius of 150. Line segment FB is perpendicular to DB, and point G is the foot of the perpendicular dropped from point B onto the line DF. Find the area of triangle BDG.
\end{lstlisting}
}
\end{tcolorbox}
\begin{tcolorbox}[breakable, enhanced jigsaw]
\small{
\begin{lstlisting}[frame=none]
Solution:
Let ABCD be a square with a side length of 100. Point F lies on a circle centered at D with a radius of 150, and FB is perpendicular to DB. Point G is the foot of the perpendicular from B to line DF.

Since triangle BDF and triangle BDG share ∠BDF and are both right triangles, they are similar. Using the similarity ratio, we have:

\[
\frac{BD}{DF} = \frac{DG}{BD}
\]

From the square, \( BD = 100\) and \( DF = 150 \), so:

\[
\frac{100}{150} = \frac{DG}{100} \implies DG = \frac{400}{3}.
\]

Since \( DF = DG + FG \) and \( DF = 150 \), substituting \( DG = \frac{400}{3} \):

\[
\frac{400}{3} + FG = 150 \implies FG = \frac{50}{3}.
\]

Triangle BDG has a right angle at G. Its area is given by:

\[
\text{Area} = \frac{1}{2} \times BG \times DG.
\]

Using the similarity ratio from triangles BDG and FBG:

\[
\frac{BG}{DG} = \frac{FG}{BG} \implies BG^2 = DG \cdot FG.
\]

Substituting \( DG = \frac{400}{3} \) and \( FG = \frac{50}{3} \):

\[
BG^2 = \frac{400}{3} \cdot \frac{50}{3} = \frac{20000}{9} \implies BG = \sqrt{\frac{20000}{9}} = \frac{100\sqrt{2}}{3}.
\]

The area of triangle BDG is:

\[
\text{Area} = \frac{1}{2} \times BG \times DG = \frac{1}{2} \times \frac{100\sqrt{2}}{3} \times \frac{400}{3} = \frac{20000 \sqrt{2}}{9}.
\]

\[
\boxed{\frac{20000 \sqrt{2}}{9}}
\]
\end{lstlisting}
}
\end{tcolorbox}

\begin{figure}[ht]
\begin{center}
\includegraphics[width=0.5\linewidth]{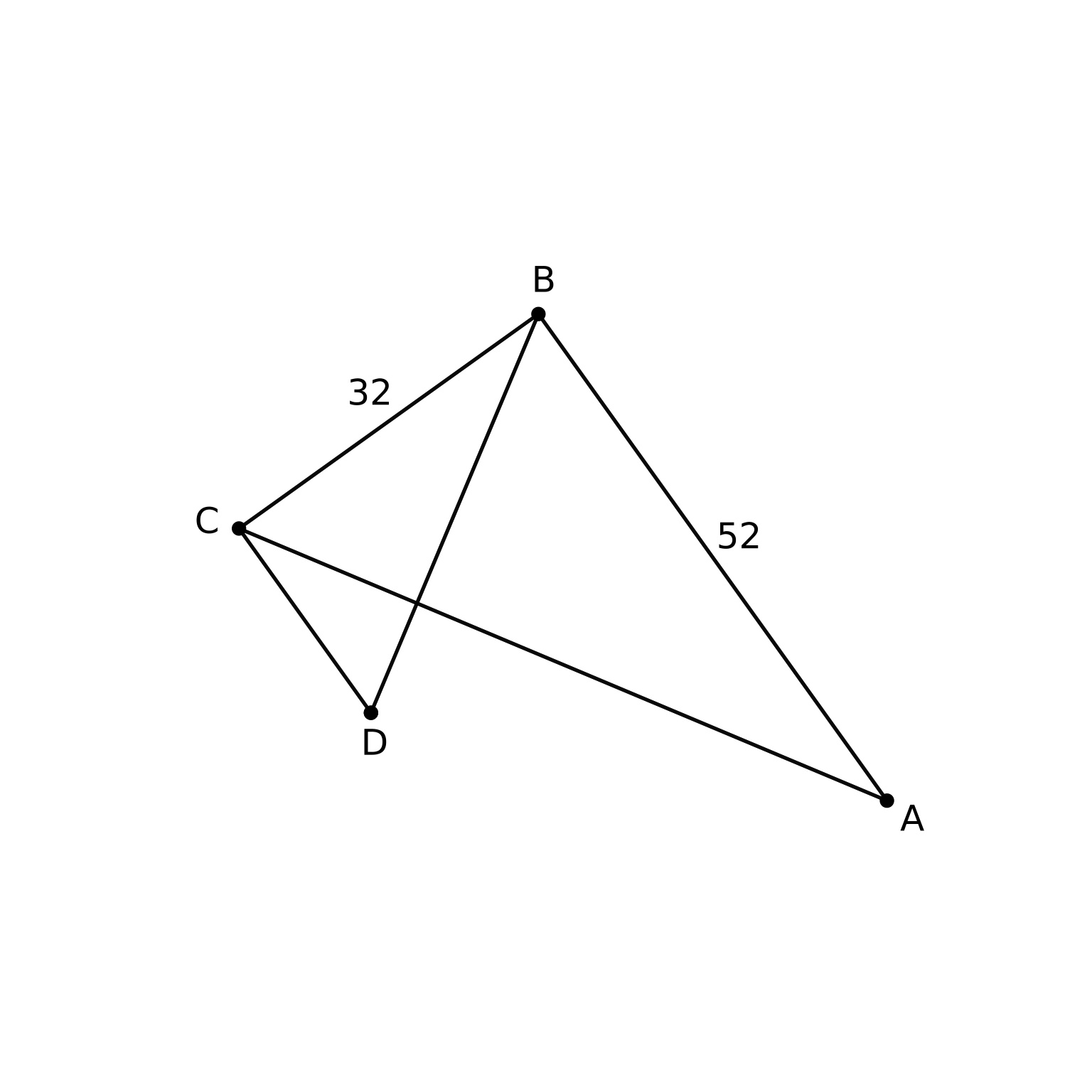}
\end{center}
\vspace{-20pt}
\caption{Diagram of a generated multiple-choice calculation problem.}
\label{fig:example2}
\end{figure}

The second example is shown below, with its corresponding diagram provided in Figure~\ref{fig:example2}.
\begin{tcolorbox}[breakable, enhanced jigsaw]
\small{
\begin{lstlisting}[frame=none]
Problem: 
In triangle ABC, ∠ABC is a right angle, AB = 52, and BC = 32. D is a point on the line passing through C and parallel to AB, such that DB is perpendicular to AC. Find the area of triangle BCD.

Choices:
A: 2048/13
B: 4096/13
C: 5120/13
D: 1024/13 
\end{lstlisting}
}
\end{tcolorbox}
\begin{tcolorbox}[breakable, enhanced jigsaw]
\small{
\begin{lstlisting}[frame=none]
Solution:
Given that \(\angle ABC\) is a right angle, \(AB\) is perpendicular to \(BC\). Since \(AB\) is parallel to \(CD\), \(CD\) is also perpendicular to \(BC\), making \(\angle BCD = 90^\circ\). The area of triangle \(BCD\) is given by \(\frac{1}{2} \cdot BC \cdot CD\).

Triangles \(ABC\) and \(BCD\) are similar because \(\angle ABC = \angle BCD\) (both \(90^\circ\)), and the other angles are congruent by the geometric constraints. By the similarity ratio, \(\frac{AB}{BC} = \frac{BC}{CD}\). Substituting the given side lengths \(AB = 52\) and \(BC = 32\), we solve for \(CD\):

\[
\frac{52}{32} = \frac{32}{CD} \implies CD = \frac{32 \cdot 32}{52} = \frac{1024}{52} = \frac{256}{13}.
\]

The area of triangle \(BCD\) is:

\[
\text{Area} = \frac{1}{2} \cdot BC \cdot CD = \frac{1}{2} \cdot 32 \cdot \frac{256}{13} = \frac{4096}{13}.
\]

\[
\boxed{B}
\]
\end{lstlisting}
}
\end{tcolorbox}

\begin{figure}[ht]
\begin{center}
\includegraphics[width=0.5\linewidth]{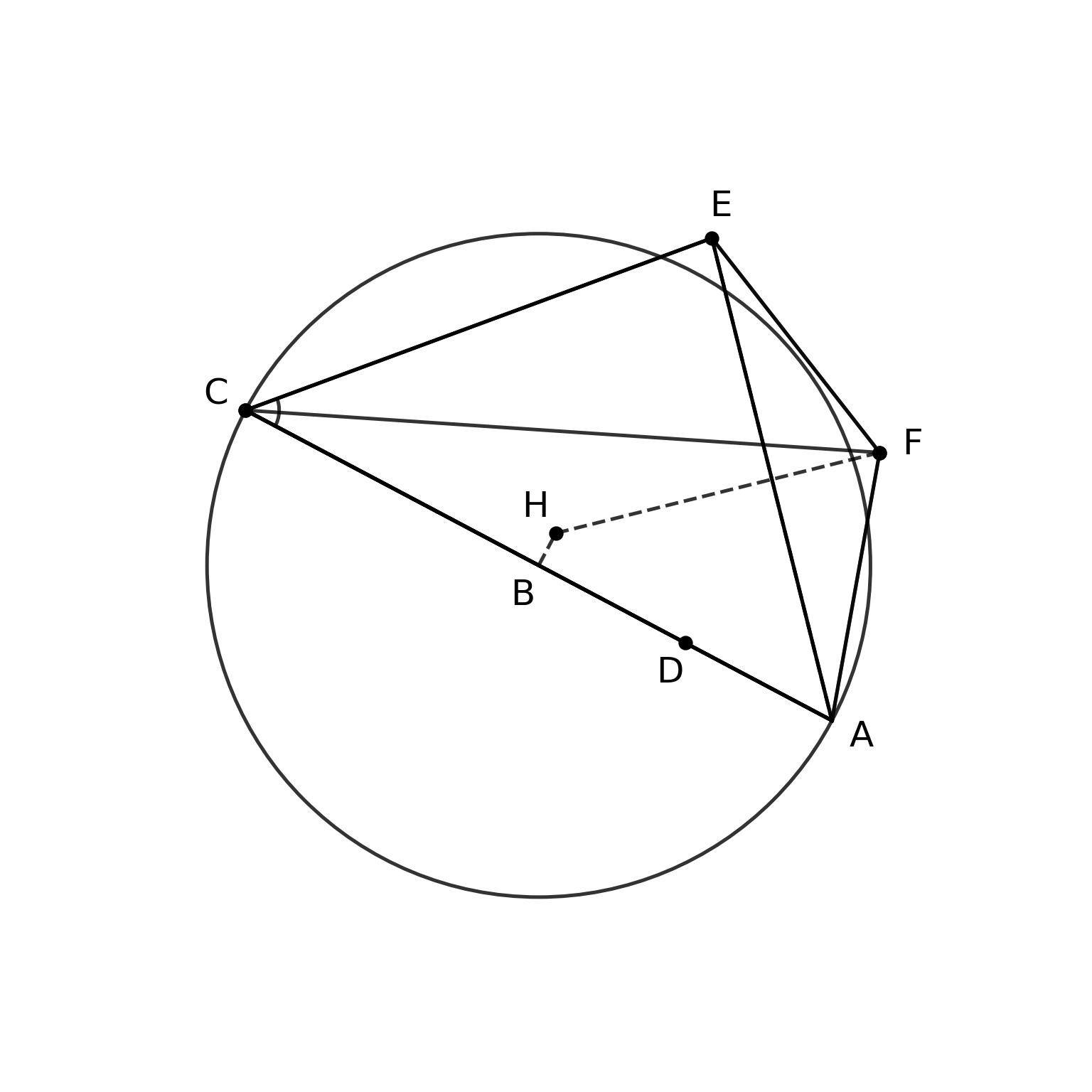}
\end{center}
\vspace{-20pt}
\caption{Diagram of a synthetic problem with its auxiliary constructions.} 
\label{fig:example3}
\end{figure}

We also include two examples of synthetic problems that require auxiliary constructions. The first example is shown below, and its associated diagram is provided in Figure~\ref{fig:example3}.
\begin{tcolorbox}[breakable, enhanced jigsaw]
\small{
\begin{lstlisting}[frame=none]
Problem:
a,b = construct_segment(), c = construct_on_circle(b,a), 
c = construct_on_line(b,a), e = construct_on_bline(c,a), 
f = construct_angle_bisector(b,c,e), f = construct_on_bline(e,a)
Goal:
Concyclic(a,c,e,f)
\end{lstlisting}
}
\end{tcolorbox}
\begin{tcolorbox}[breakable, enhanced jigsaw]
\small{
\begin{lstlisting}[frame=none]
Auxiliary Constructions:
d = construct_midpoint(b,a), 
h = construct_intersection_tt(f,a,e,b,c,d)
\end{lstlisting}
}
\end{tcolorbox}

\begin{figure}[ht]
\begin{center}
\includegraphics[width=0.5\linewidth]{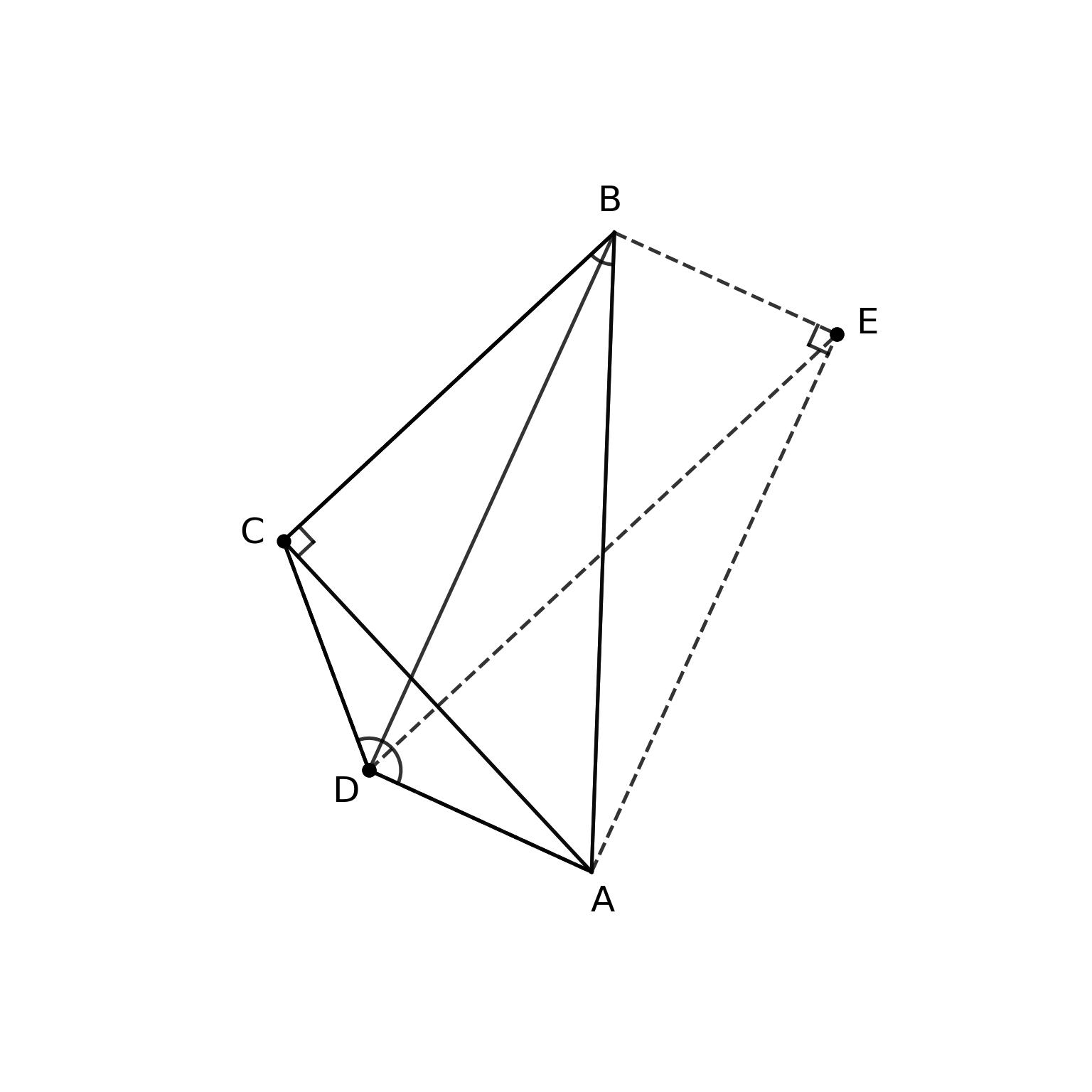}
\end{center}
\vspace{-20pt}
\caption{Diagram of a synthetic problem with its auxiliary constructions.} 
\label{fig:example4}
\end{figure}

The second example is presented below, with its corresponding diagram shown in Figure~\ref{fig:example4}.
\begin{tcolorbox}[breakable, enhanced jigsaw]
\small{
\begin{lstlisting}[frame=none]
Problem:
a,b = construct_segment(), c = construct_on_dia(b,a), 
d = construct_on_bline(c,a), d = construct_angle_bisector(c,b,a)
Goal:
Angle_a_b_c + Angle_a_d_c - 180
\end{lstlisting}
}
\end{tcolorbox}
\begin{tcolorbox}[breakable, enhanced jigsaw]
\small{
\begin{lstlisting}[frame=none]
Auxiliary Constructions:
e = construct_on_dia(a,b), e = construct_angle_bisector(c,d,a)
\end{lstlisting}
}
\end{tcolorbox}

\section{Experiments}
\subsection{Examples of Generated Formal Solutions}
\label{app:proof}
We compare the proofs generated by \solver{} against those produced by AlphaGeometry~\citep{alphageometry}, Newclid~\citep{newclid}, and PyEuclid~\citep{pyeuclid}. For this evaluation, we randomly selected two problems: one from JGEX-AG-231~\citep{alphageometry} and another from IMO-AG-30~\citep{alphageometry}.

\begin{figure}[ht]
\begin{center}
\includegraphics[width=0.5\linewidth]{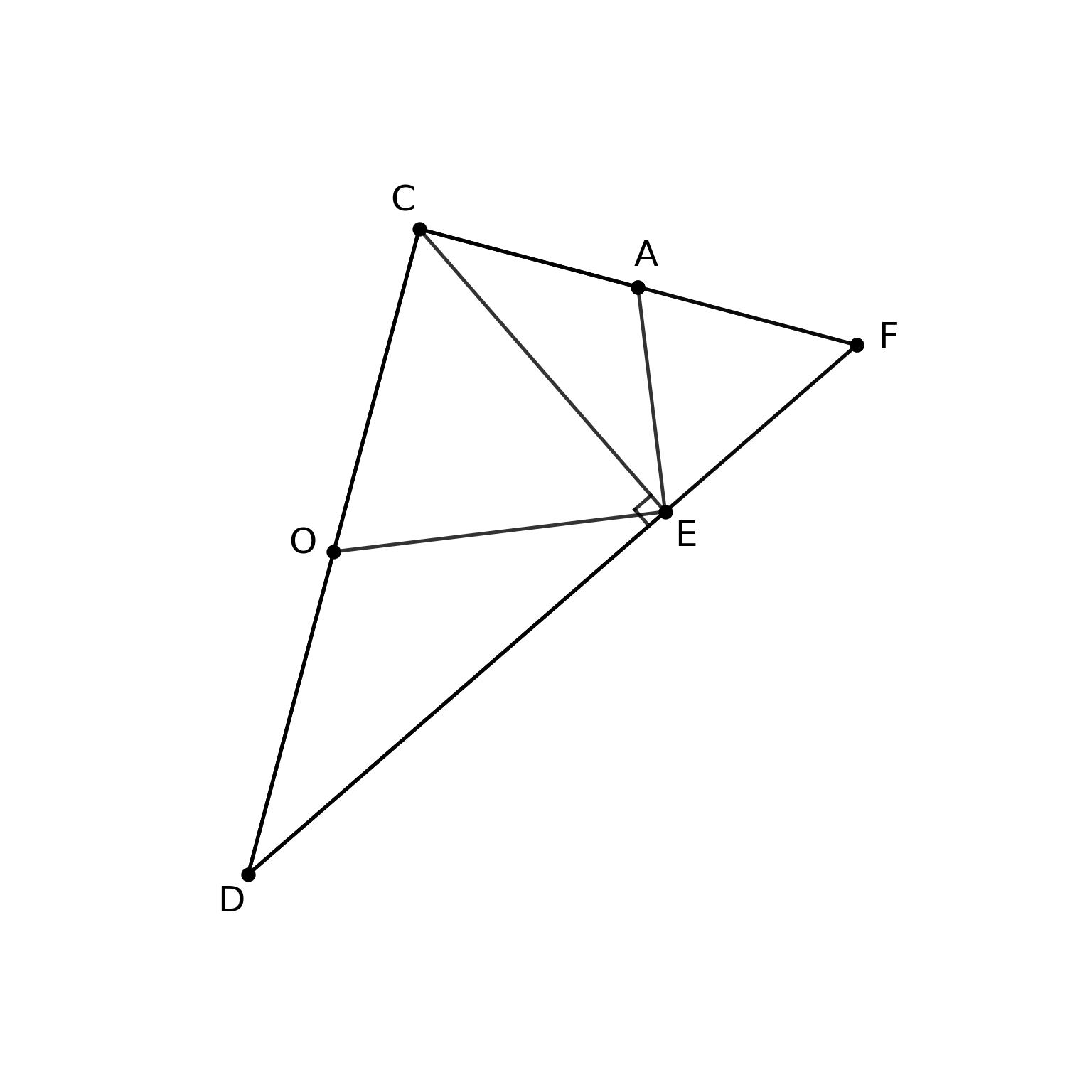}
\end{center}
\vspace{-10pt}
\caption{Diagram of a geometry problem selected from JGEX-AG-231.} 
\label{fig:jgex}
\end{figure}

The natural language formulation of the first problem is given below, and the corresponding diagram is shown in Figure~\ref{fig:jgex}.
\begin{tcolorbox}[breakable, enhanced jigsaw]
\small{
\begin{lstlisting}[frame=none]
In triangle ECD, ∠E is a right angle. O is the midpoint of side DC. Line AC is perpendicular to side DC, and AE is perpendicular to EO. Line CA intersects at a point F, and line DE also passes through point F. Prove that AE is equal to AF.
\end{lstlisting}
}
\end{tcolorbox}

The proof produced by \solver{} for this problem is shown below:
\begin{tcolorbox}[breakable, enhanced jigsaw]
\small{
\begin{lstlisting}[frame=none]
Solution:
1. Collinear(d,e,f) => Angle_e_d_o - Angle_f_d_o & Angle_a_f_d - Angle_a_f_e & Angle_d_e_o + Angle_f_e_o - 180
2. Collinear(c,d,o) => Parallel(c,d,d,o)
3. Perpendicular(a,c,c,d) & Parallel(c,d,d,o) => Perpendicular(a,c,d,o)
4. Collinear(a,c,f) => Parallel(a,c,a,f)
5. Perpendicular(a,c,d,o) & Parallel(a,c,a,f) => Perpendicular(a,f,d,o)
6. Perpendicular(a,f,d,o) => Angle_a_f_d + Angle_f_d_o - 90
7. Angle_e_d_o - Angle_f_d_o & Angle_a_f_d - Angle_a_f_e & Angle_a_f_d + Angle_f_d_o - 90 => Angle_a_f_e + Angle_e_d_o - 90
8. Perpendicular(a,e,e,o) => Angle_a_e_o - 90
9. Angle_a_e_f + Angle_a_e_o - Angle_f_e_o & Angle_d_e_o + Angle_f_e_o - 180 & Angle_a_e_o - 90 => Angle_a_e_f + Angle_d_e_o - 90
10. Midpoint(o,c,d) & Perpendicular(c,e,d,e) => Length_d_o - Length_e_o
11. Length_d_o - Length_e_o => Angle_d_e_o - Angle_e_d_o
12. Angle_a_f_e + Angle_e_d_o - 90 & Angle_a_e_f + Angle_d_e_o - 90 & Angle_d_e_o - Angle_e_d_o => Angle_a_e_f - Angle_a_f_e
13. Angle_a_e_f - Angle_a_f_e => Length_a_e - Length_a_f

\end{lstlisting}
}
\end{tcolorbox}

The proof produced by AlphaGeometry for this problem is shown below:
\begin{tcolorbox}[breakable, enhanced jigsaw]
\small{
\begin{lstlisting}[frame=none]
 * Proof steps:
001. C,O,D are collinear [01] & CD ⟂ AC [03] ⇒  CO ⟂ CA [07]
002. CO ⟂ CA [07] & AE ⟂ EO [04] ⇒  (*@\textcolor{red}{$\angle$COE = $\angle$CAE}@*) [08]
003. C,A,F are collinear [05] & C,O,D are collinear [01] & (*@\textcolor{red}{$\angle$COE = $\angle$CAE}@*) [08] ⇒  ∠FAE = ∠COE [09]
004. D,F,E are collinear [06] & DE ⟂ CE [00] ⇒  FD ⟂ CE [10]
005. AE ⟂ EO [04] & FD ⟂ CE [10] ⇒  (*@\textcolor{blue}{$\angle$(AE-FD)}@*) = ∠OEC [11]
006. D,F,E are collinear [06] & (*@\textcolor{blue}{$\angle$(AE-FD)}@*) = ∠OEC [11] ⇒  ∠FEA = ∠CEO [12]
007. ∠FAE = ∠COE [09] & ∠FEA = ∠CEO [12] (Similar Triangles)⇒  OC:OE = AF:AE [13]
008. C,O,D are collinear [01] & OD = OC [02] ⇒  O is midpoint of DC [14]
009. CE ⟂ DE [00] & O is midpoint of DC [14] ⇒  CO = EO [15]
010. OC:OE = AF:AE [13] & CO = EO [15] ⇒  AF = AE

\end{lstlisting}
}
\end{tcolorbox}

The proof produced by Newclid for this problem is shown below:
\begin{tcolorbox}[breakable, enhanced jigsaw]
\small{
\begin{lstlisting}[frame=none]
# Proof:
000. | O is the midpoint of CD [C0], CE ⟂ DE [C1] =(r19 Hypotenuse is diameter)> CO = EO [0]
001. | CO = EO [0] =(r13 Isosceles triangle equal angles)> ∠(CE,CO) = ∠(EO,CE) [1]
002. | A, C, F are collinear [C2], A ≠ C [N0], A ≠ F [N1], C ≠ F [N2] =(r82 Parallel from collinear)> AF ∥ AC [2]
003. | O is the midpoint of CD [C0] =(r56 Properties of midpoint (coll))> C, D, O are collinear [3]
004. | C, D, O are collinear [3], C ≠ D [N3], C ≠ O [N4], D ≠ O [N5] =(r82 Parallel from collinear)> CO ∥ CD [4]
005. | D, E, F are collinear [C3], D ≠ E [N6], D ≠ F [N7], E ≠ F [N8] =(r82 Parallel from collinear)> DE ∥ EF [5]
006. | ∠(CE,CO) = ∠(EO,CE) [1], AF ∥ AC [2], CO ∥ CD [4], DE ∥ EF [5], AC ⟂ CD [C4], AE ⟂ EO [C5], CE ⟂ DE [C1] =(AR Deduction)> (*@\textcolor{blue}{$\angle$(AE,EF)}@*) = (*@\textcolor{blue}{$\angle$(EF,AF)}@*) [6]
007. | (*@\textcolor{blue}{$\angle$(AE,EF)}@*) = (*@\textcolor{blue}{$\angle$(EF,AF)}@*) [6], A, E, F are not collinear [N9] =(r14 Equal base angles imply isosceles)> AF = AE [7]
\end{lstlisting}
}
\end{tcolorbox}

The proof produced by PyEuclid for this problem is shown below:
\begin{tcolorbox}[breakable, enhanced jigsaw]
\small{
\begin{lstlisting}[frame=none]
* Proof steps:
001. Length_c_o - Length_d_o &
Collinear(c,d,o) &
Between(o,c,d) ⇒ -Length_c_d/2 + Length_d_o
002. Angle_c_d_e - Angle_f_d_o &
Angle_c_d_f - Angle_f_d_o ⇒ Angle_c_d_e - Angle_c_d_f
003. Perpendicular(a,c,c,d) &
Parallel(a,c,c,f) ⇒ Angle_d_c_f - pi/2
004. Angle_d_c_f - pi/2(3) &
Angle_c_e_d - pi/2 ⇒ Angle_c_e_d - Angle_d_c_f
005. Not(Collinear(c,d,e)) &
Angle_c_d_e - Angle_c_d_f(2) &
Angle_c_e_d - Angle_d_c_f(4) ⇒ Length_c_d/Length_d_f - Length_d_e/Length_c_d
006. Perpendicular(c,e,d,e) &
Parallel(d,e,e,f) ⇒ Angle_c_e_f - pi/2
007. Angle_c_e_d - pi/2 &
Angle_c_e_f - pi/2(6) ⇒ -Angle_c_e_d + Angle_c_e_f
008. Angle_c_e_d - pi/2 &
Angle_c_d_e - Angle_f_d_o &
Angle_c_d_e + Angle_c_e_d + Angle_d_c_e - pi ⇒ Angle_d_c_e + Angle_f_d_o - pi/2
009. Angle_c_d_f - Angle_f_d_o &
Angle_c_f_d - Angle_c_f_e &
Angle_c_d_f + Angle_c_f_d + Angle_d_c_f - pi &
Angle_d_c_f - pi/2 ⇒ Angle_c_f_e + Angle_f_d_o - pi/2
010. Angle_d_c_e + Angle_f_d_o - pi/2(8) &
Angle_c_f_e + Angle_f_d_o - pi/2(9) ⇒ Angle_c_f_e - Angle_d_c_e
011. Not(Collinear(c,e,f)) &
-Angle_c_e_d + Angle_c_e_f(7) &
Angle_c_f_e - Angle_d_c_e(10) ⇒ Length_c_e/Length_d_e - Length_e_f/Length_c_e
012. Angle_a_e_o - pi/2 &
Angle_a_f_e - Angle_c_f_d &
-Angle_c_d_f + Angle_f_d_o &
-Angle_a_e_f - Angle_a_e_o + Angle_f_e_o &
Angle_a_e_f + Angle_a_f_e + Angle_e_a_f - pi &
Angle_c_d_f + Angle_c_f_d + Angle_d_c_f - pi &
Angle_d_c_f - pi/2 ⇒ Angle_e_a_f - Angle_f_d_o + Angle_f_e_o - pi
013. -Angle_e_d_o + Angle_f_d_o &
Angle_d_e_o + Angle_d_o_e + Angle_e_d_o - pi &
Angle_c_o_e + Angle_d_o_e - pi &
Angle_d_e_o + Angle_f_e_o - pi ⇒ Angle_c_o_e - Angle_f_d_o + Angle_f_e_o - pi
014. Angle_e_a_f - Angle_f_d_o + Angle_f_e_o - pi(12) &
Angle_c_o_e - Angle_f_d_o + Angle_f_e_o - pi(13) ⇒ -Angle_c_o_e + Angle_e_a_f
015. -Angle_c_e_f - Angle_c_e_o + Angle_f_e_o &
Angle_c_e_f - pi/2 ⇒ Angle_c_e_o - Angle_f_e_o + pi/2
016. Angle_a_e_o - pi/2 &
-Angle_a_e_f - Angle_a_e_o + Angle_f_e_o ⇒ Angle_a_e_f - Angle_f_e_o + pi/2
017. Angle_c_e_o - Angle_f_e_o + pi/2(15) &
Angle_a_e_f - Angle_f_e_o + pi/2(16) ⇒ Angle_a_e_f - Angle_c_e_o
018. Not(Collinear(a,e,f)) &
-Angle_c_o_e + Angle_e_a_f(14) &
Angle_a_e_f - Angle_c_e_o(17) ⇒ Length_a_f/Length_c_o - Length_e_f/Length_c_e
019. -Length_c_o + Length_d_o &
-Length_c_d/2 + Length_d_o(1) &
Length_c_d/Length_d_f - Length_d_e/Length_c_d(5) &
Length_c_e/Length_d_e - Length_e_f/Length_c_e(11) &
Length_a_f/Length_c_o - Length_e_f/Length_c_e(18) ⇒ Length_a_f - sqrt(Length_d_f)*sqrt(Length_e_f)/2
020. -Angle_a_f_e + Angle_c_f_d &
Angle_c_d_f - Angle_f_d_o &
Angle_c_d_f + Angle_c_f_d + Angle_d_c_f - pi &
Angle_d_c_f - pi/2 ⇒ Angle_a_f_e + Angle_f_d_o - pi/2
021. Angle_c_e_d - pi/2 &
Angle_c_d_e - Angle_f_d_o &
Angle_d_c_e - Angle_e_c_o &
Angle_c_d_e + Angle_c_e_d + Angle_d_c_e - pi ⇒ Angle_e_c_o + Angle_f_d_o - pi/2
022. Angle_a_f_e + Angle_f_d_o - pi/2(20) &
Angle_e_c_o + Angle_f_d_o - pi/2(21) ⇒ Angle_a_f_e - Angle_e_c_o
023. Not(Collinear(a,e,f)) &
Angle_a_e_f - Angle_c_e_o(17) &
Angle_a_f_e - Angle_e_c_o(22) ⇒ Length_a_e/Length_e_o - Length_e_f/Length_c_e
024. Angle_e_d_o - Angle_f_d_o &
Angle_d_e_o + Angle_d_o_e + Angle_e_d_o - pi &
Angle_d_e_o + Angle_f_e_o - pi ⇒ Angle_d_o_e + Angle_f_d_o - Angle_f_e_o
025. Angle_a_e_o - pi/2 &
-Angle_a_f_e + Angle_c_f_d &
Angle_c_d_f - Angle_f_d_o &
-Angle_a_e_f - Angle_a_e_o + Angle_f_e_o &
Angle_a_e_f + Angle_a_f_e + Angle_e_a_f - pi &
Angle_c_d_f + Angle_c_f_d + Angle_d_c_f - pi &
Angle_c_a_e + Angle_e_a_f - pi &
Angle_d_c_f - pi/2 ⇒ Angle_c_a_e + Angle_f_d_o - Angle_f_e_o
026. Angle_d_o_e + Angle_f_d_o - Angle_f_e_o(24) &
Angle_c_a_e + Angle_f_d_o - Angle_f_e_o(25) ⇒ Angle_c_a_e - Angle_d_o_e
027. Angle_a_c_d - pi/2 &
Angle_c_e_d - pi/2 &
-Angle_c_d_e + Angle_f_d_o &
Angle_a_c_d - Angle_a_c_e - Angle_d_c_e &
Angle_c_d_e + Angle_c_e_d + Angle_d_c_e - pi ⇒ Angle_a_c_e - Angle_f_d_o
028. Angle_a_c_e - Angle_f_d_o(27) &
Angle_e_d_o - Angle_f_d_o ⇒ Angle_a_c_e - Angle_e_d_o
029. Not(Collinear(a,c,e)) &
Angle_c_a_e - Angle_d_o_e(26) &
Angle_a_c_e - Angle_e_d_o(28) ⇒ Length_a_e/Length_e_o - Length_c_e/Length_d_e
030. Length_c_o - Length_d_o &
Collinear(c,d,o) &
Between(o,c,d) &
Perpendicular(c,e,d,e) ⇒ Length_c_o - Length_e_o
031. -Length_c_d/2 + Length_d_o(1) &
Length_a_e/Length_e_o - Length_e_f/Length_c_e(23) &
Length_c_d/Length_d_f - Length_d_e/Length_c_d(5) &
Length_a_e/Length_e_o - Length_c_e/Length_d_e(29) &
Length_d_o/Length_e_o - Length_e_o/Length_d_o ⇒ Length_a_e - sqrt(Length_d_f)*sqrt(Length_e_f)/2
032. Length_a_f - sqrt(Length_d_f)*sqrt(Length_e_f)/2(19) &
Length_a_e - sqrt(Length_d_f)*sqrt(Length_e_f)/2(31) ⇒ Length_a_e - Length_a_f
\end{lstlisting}
}
\end{tcolorbox}

\begin{figure}[ht]
\begin{center}
\includegraphics[width=0.5\linewidth]{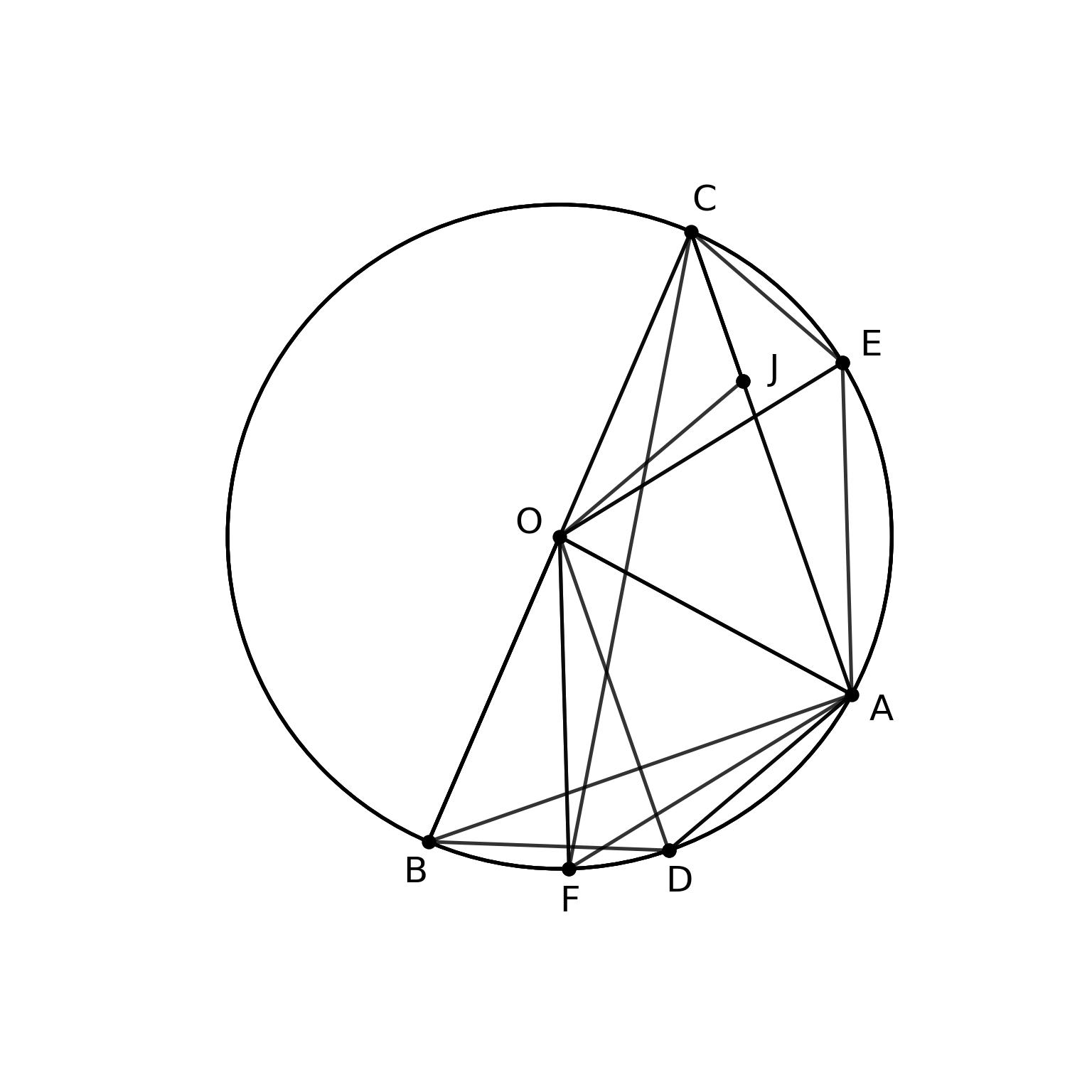}
\end{center}
\vspace{-10pt}
\caption{Diagram of a geometry problem selected from IMO-AG-30.} 
\label{fig:imo2002}
\end{figure}

For the second problem, the natural language formulation is provided below, along with its corresponding diagram in Figure~\ref{fig:imo2002}.
\begin{tcolorbox}[breakable, enhanced jigsaw]
\small{
\begin{lstlisting}[frame=none]
Let BC be a line segment, and O be the midpoint of BC. Point A lies on a circle with center O and radius OB. Point D lies on the perpendicular bisector of line segment AB and is located on the same circle with center O and radius OB. Point E is located on the perpendicular bisector of line segment OA and also lies on the circle with center O and radius OB. Similarly, point F is located on the perpendicular bisector of line segment OA and lies on the circle with center O and radius OB. Line JO is parallel to AD, and point J lies on line AC. Prove that ∠ECJ = ∠FCJ.
\end{lstlisting}
}
\end{tcolorbox}

The proof produced by \solver{} for this problem is shown below:
\begin{tcolorbox}[breakable, enhanced jigsaw]
\small{
\begin{lstlisting}[frame=none]
Solution:
1. Length_b_o - Length_c_o & Length_a_o - Length_b_o => Length_a_o - Length_c_o
2. Length_b_o - Length_f_o & Length_b_o - Length_c_o => Length_c_o - Length_f_o
3. -Length_a_o + Length_c_o & Length_c_o - Length_f_o => Angle_a_c_f - Angle_a_o_f/2
4. Collinear(a,c,j) => Angle_a_c_f - Angle_f_c_j & Angle_a_c_e - Angle_e_c_j
5. Length_b_o - Length_f_o & Length_b_o - Length_e_o => Length_e_o - Length_f_o
6. Length_a_e - Length_e_o & Length_e_o - Length_f_o & -Length_a_f + Length_f_o => Rhombus(a,e,o,f)
7. Rhombus(a,e,o,f) => Angle_a_o_e - Angle_a_o_f
8. Length_b_o - Length_c_o & Length_b_o - Length_e_o => Length_c_o - Length_e_o
9. -Length_a_o + Length_c_o & Length_c_o - Length_e_o => Angle_a_c_e - Angle_a_o_e/2
10. Angle_a_c_f - Angle_a_o_f/2 & Angle_a_c_f - Angle_f_c_j & Angle_a_o_e - Angle_a_o_f & Angle_a_c_e - Angle_a_o_e/2 & Angle_a_c_e - Angle_e_c_j => Angle_e_c_j - Angle_f_c_j

\end{lstlisting}
}
\end{tcolorbox}

The proof produced by AlphaGeometry for this problem is shown below:
\begin{tcolorbox}[breakable, enhanced jigsaw]
\small{
\begin{lstlisting}[frame=none]
 * Proof steps:
001. OE = OB [03] & OF = OB [05] & OA = OB [01] & OD = OB [02] & OB = OC [00] ⇒  E,A,F,C are concyclic [08]
002. E,A,F,C are concyclic [08] ⇒  ∠AEF = ∠ACF [09]
003. E,A,F,C are concyclic [08] ⇒  ∠EFA = ∠ECA [10]
004. EO = EA [04] & OE = OB [03] & OF = OB [05] & FO = FA [06] ⇒  AF = AE [11]
005. AF = AE [11] ⇒  ∠EFA = ∠AEF [12]
006. J,A,C are collinear [07] & ∠AEF = ∠ACF [09] & ∠EFA = ∠AEF [12] & ∠EFA = ∠ECA [10] ⇒  ∠ECJ = ∠JCF
\end{lstlisting}
}
\end{tcolorbox}

The proof produced by Newclid for this problem is shown below:
\begin{tcolorbox}[breakable, enhanced jigsaw]
\small{
\begin{lstlisting}[frame=none]
# Proof:
000. | O is the midpoint of BC [C1] =(r51 Midpoint splits in two)> BC:BO = 2/1 [0]
001. | O is the midpoint of BC [C1] =(r51 Midpoint splits in two)> BC:CO = 2/1 [1]
002. | AO = BO [C0], BC:BO = 2/1 [0], BC:CO = 2/1 [1] =(AR Deduction)> CO = AO [2]
003. | CO = AO [2] =(r13 Isosceles triangle equal angles)> (*@\textcolor{blue}{$\angle$(AC,AO)}@*) = (*@\textcolor{blue}{$\angle$(CO,AC)}@*) [3]
004. | (*@\textcolor{blue}{$\angle$(AE,AO)}@*) = (*@\textcolor{blue}{$\angle$(AO,EO)}@*) [C3], (*@\textcolor{blue}{$\angle$(AF,AO)}@*) = (*@\textcolor{blue}{$\angle$(AO,FO)}@*) [C2] =(AR Deduction)> (*@\textcolor{blue}{$\angle$(AE,AF)}@*) = (*@\textcolor{blue}{$\angle$(FO,EO)}@*) [4]
005. | AE = EO [C4], EO = BO [C5], FO = AF [C6], FO = BO [C7] =(AR Deduction)> AE:AF = FO:EO [5]
006. | (*@\textcolor{blue}{$\angle$(AE,AF)}@*) = (*@\textcolor{blue}{$\angle$(FO,EO)}@*) [4], AE:AF = FO:EO [5], ▲AEF has the same orientation as ▲EOF [N0] =(r62 SAS Similarity of triangles (Direct))> ▲AEF ≅ ▲OFE [6]
007. | ▲AEF has the same orientation as ▲EOF [N0], ▲AEF ≅ ▲OFE [6] =(r52 Properties of similar triangles (Direct))> (*@\textcolor{blue}{$\angle$(AF,EF)}@*) = (*@\textcolor{blue}{$\angle$(EO,EF)}@*) [7]
008. | EO = BO [C5], BC:BO = 2/1 [0], BC:CO = 2/1 [1] =(AR Deduction)> EO = CO [8]
009. | EO = CO [8] =(r13 Isosceles triangle equal angles)> (*@\textcolor{blue}{$\angle$(CE,CO)}@*) = (*@\textcolor{blue}{$\angle$(EO,CE)}@*) [9]
010. | FO = BO [C7], BC:BO = 2/1 [0], BC:CO = 2/1 [1] =(AR Deduction)> CO = FO [10]
011. | CO = FO [10] =(r13 Isosceles triangle equal angles)> (*@\textcolor{blue}{$\angle$(CF,CO)}@*) = (*@\textcolor{blue}{$\angle$(FO,CF)}@*) [11]
012. | A, C, J are collinear [C8], A ≠ C [N1], A ≠ J [N2], C ≠ J [N3] =(r82 Parallel from collinear)> CJ ∥ AC [12]
013. | (*@\textcolor{blue}{$\angle$(AC,AO)}@*) = (*@\textcolor{blue}{$\angle$(CO,AC)}@*) [3], (*@\textcolor{blue}{$\angle$(AF,AO)}@*) = (*@\textcolor{blue}{$\angle$(AO,FO)}@*) [C2], (*@\textcolor{blue}{$\angle$(AF,EF)}@*) = (*@\textcolor{blue}{$\angle$(EO,EF)}@*) [7], (*@\textcolor{blue}{$\angle$(CE,CO)}@*) = (*@\textcolor{blue}{$\angle$(EO,CE)}@*) [9], (*@\textcolor{blue}{$\angle$(CF,CO)}@*) = (*@\textcolor{blue}{$\angle$(FO,CF)}@*) [11], CJ ∥ AC [12] =(AR Deduction)> (*@\textcolor{blue}{$\angle$(CE,CJ)}@*) = (*@\textcolor{blue}{$\angle$(CJ,CF)}@*) [13]
\end{lstlisting}
}
\end{tcolorbox}

The proof produced by PyEuclid for this problem is shown below:
\begin{tcolorbox}[breakable, enhanced jigsaw]
\small{
\begin{lstlisting}[frame=none]
* Proof steps:
001. Length_a_o - Length_b_o &
-Length_b_o + Length_f_o ⇒ Length_a_o - Length_f_o
002. -Length_a_e + Length_e_o &
-Length_b_o + Length_e_o &
-Length_b_o + Length_f_o ⇒ Length_a_e - Length_f_o
003. Length_a_o - Length_f_o(1) &
Length_a_e - Length_f_o(2) ⇒ Length_a_e - Length_a_o
004. Not(Collinear(a,e,o)) &
Length_a_e - Length_a_o(3) ⇒ Angle_a_e_o - Angle_a_o_e
005. -Length_b_o + Length_e_o &
-Length_b_o + Length_f_o ⇒ Length_e_o - Length_f_o
006. Not(Collinear(a,e,o)) &
Length_a_e - Length_e_o ⇒ -Angle_a_o_e + Angle_e_a_o
007. Length_b_o - Length_c_o &
-Length_b_o + Length_f_o ⇒ Length_c_o - Length_f_o
008. Length_a_o - Length_f_o(1) &
Length_c_o - Length_f_o(7) ⇒ Length_a_o - Length_c_o
009. Length_e_o - Length_f_o(5) &
Length_c_o - Length_f_o(7) ⇒ Length_c_o - Length_e_o
010. SameSide(c,o,a,e) &
Length_a_o - Length_c_o(8) &
Length_c_o - Length_e_o(9) ⇒ Angle_a_c_e - Angle_a_o_e/2
011. Angle_a_c_e - Angle_e_c_j &
Angle_a_e_o + Angle_a_o_e + Angle_e_a_o - pi &
Angle_a_e_o - Angle_a_o_e(4) &
-Angle_a_o_e + Angle_e_a_o(6) &
Angle_a_c_e - Angle_a_o_e/2(10) ⇒ Angle_e_c_j - pi/6
012. Length_a_o - Length_f_o(1) &
Length_a_f - Length_f_o ⇒ Length_a_f - Length_a_o
013. Not(Collinear(a,f,o)) &
Length_a_f - Length_a_o(12) ⇒ Angle_a_f_o - Angle_a_o_f
014. Not(Collinear(a,f,o)) &
Length_a_f - Length_f_o ⇒ -Angle_a_o_f + Angle_f_a_o
015. SameSide(c,o,a,f) &
Length_a_o - Length_c_o(8) &
Length_c_o - Length_f_o ⇒ Angle_a_c_f - Angle_a_o_f/2
016. Angle_a_c_f - Angle_f_c_j &
Angle_a_f_o + Angle_a_o_f + Angle_f_a_o - pi &
Angle_a_f_o - Angle_a_o_f(13) &
-Angle_a_o_f + Angle_f_a_o(14) &
Angle_a_c_f - Angle_a_o_f/2(15) ⇒ Angle_f_c_j - pi/6
017. Angle_e_c_j - pi/6(11) &
Angle_f_c_j - pi/6(16) ⇒ Angle_e_c_j - Angle_f_c_j
\end{lstlisting}
}
\end{tcolorbox}

It is important to note that proofs generated by different systems can vary substantially in both style and strategy. In contrast to \solver{}, AlphaGeometry and Newclid rely on full-angle formalization~\citep{full-angle}, which often fails to distinguish an angle from its supplement and may therefore yield ``incorrect'' angle relations compared to human reasoning (highlighted in red). Furthermore, these systems represent angles as pairs of lines (shown in blue), which introduces additional ambiguity. By comparison, the proofs generated by \solver{} faithfully capture the angle relations shown in the diagram, yielding a more human-like representation. Compared to PyEuclid, \solver{} produces proofs that are about twice as compact and avoid redundant equations, making them significantly clearer and easier for humans to read.

\subsection{Ablation Study over Training Datasets}
\label{app:mixed_datasets}
To perform a controlled ablation, we fine-tune Qwen2.5-VL-7B~\citep{qwen2.5vl} on several open-source datasets under matched training settings. We consider Geo170K~\citep{g-llava}, GeoGen~\citep{geogen}, and TR-CoT~\citep{r-cot}. Concretely, we train models on 20K subsets of Geo170K, GeoGen, and TR-CoT, as well as two 20K mixtures (10K Geo170K + 10K GeoGen and 10K Geo170K + 10K TR-CoT). We additionally fine-tune on Ours 10K as an ablation.

\begin{table}[h]
\centering
\caption{Accuracy (\%) on four benchmarks under different training data choices.}
\label{tab:dataset_ablation}
\vspace{-0.5em}
\begin{tabular}{lcccc}
\toprule
Training Data & GeoQA & Geometry3K & MathVista & MathVerse \\
\midrule
Qwen2.5-VL-7B (no fine-tuning)    & 69.4 & 56.4 & 72.2 & 44.1 \\
\midrule
Geo170K 20K                 & 75.0 & 57.8 & 70.1 & 50.4 \\
GeoGen 20K                  & 73.1 & 45.1 & 63.6 & 40.2 \\
TR-CoT 20K                  & 75.0 & 57.3 & 72.1 & 47.8 \\
Geo170K 10K + GeoGen 10K    & 75.3 & 59.8 & 62.9 & 44.9 \\
Geo170K 10K + TR-CoT 10K    & 72.8 & 55.2 & 74.0 & 44.5 \\
Ours 10K                    & 74.7 & 63.6 & 72.1 & 50.4 \\
\midrule
Geo170K 10K + Ours 10K      & 76.6 & 61.0 & 74.7 & 51.0 \\
\bottomrule
\end{tabular}
\vspace{-0.75em}
\end{table}

Table~\ref{tab:dataset_ablation} show that, under comparable compute, Euclid-Omni data consistently yields competitive or superior performance, especially on the Geometry3K dataset. This demonstrates that high-fidelity, symbolically verified constructions, rather than simply increasing dataset size, are essential for improving VLM performance on geometry reasoning benchmarks.

\subsection{Examples of Generated Natural Language Solutions}
\label{app:informal_calc}
We provide two examples from the GeoQA~\citep{chen2021geoqa} and Geometry3K~\citep{inter-gps} datasets and compare the solutions generated by our finetuned model with those produced by the base model Qwen2.5-VL~\citep{qwen2.5vl}. The results illustrate that training on our synthetic dataset improves the model’s ability to understand and solve informal geometry problems.

\begin{figure}[ht]
\begin{center}
\includegraphics[width=0.25\linewidth]{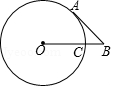}
\end{center}
\vspace{-10pt}
\caption{Diagram of a problem selected from GeoQA.} 
\label{fig:geoqa_272}
\end{figure}

The natural language description of the problem from the GeoQA dataset is presented below, with the corresponding diagram shown in Figure~\ref{fig:geoqa_272}. This example demonstrates that our model correctly interprets diagrammatic relationships through finetuning.
\begin{tcolorbox}[breakable, enhanced jigsaw]
\small{
\begin{lstlisting}[frame=none]
As shown in the figure, it is known that BA is the tangent of circle O, and connect OB to intersect circle O at point C. If angle B = 45.0 and the length of AB is 2.0, then the length of BC is ()
Choices:
A:2\sqrt{2}-1
B:\sqrt{2}
C:2\sqrt{2}-2
D:2-\sqrt{2}
\end{lstlisting}
}
\end{tcolorbox}

The solution produced by our model for this problem is shown below:
\begin{tcolorbox}[breakable, enhanced jigsaw]
\small{
\begin{lstlisting}[frame=none]
Since BA is a tangent line to circle O, we have OA perpendicular to AB. Given that angle B is 45 degrees, we can find that OA equals AB, which is 2. In right triangle AOB, using the Pythagorean theorem, we can find OB as \sqrt{OA^2+AB^2}, which equals 2\sqrt{2}. Therefore, BC equals OB minus OC, which is 2\sqrt{2}-2. Hence, the answer is \boxed{C}
\end{lstlisting}
}
\end{tcolorbox}

The solution produced by the base model for this problem is shown below:
\begin{tcolorbox}[breakable, enhanced jigsaw]
\small{
\begin{lstlisting}[frame=none]
Since AB is a tangent line to circle O, we have OA perpendicular  AB. In right triangle AOB, we know that angle B is 45° and AB is 2. Therefore, OA = OB = AB * sin(45°) = \sqrt{2}. Hence, OC = OA = \sqrt{2}. Thus, BC = OB - OC = (\sqrt{2}) - (\sqrt{2})/2 = (\sqrt{2})/2. Therefore, the answer is option D.
Answer:D
\end{lstlisting}
}
\end{tcolorbox}

\begin{figure}[ht]
\begin{center}
\includegraphics[width=0.4\linewidth]{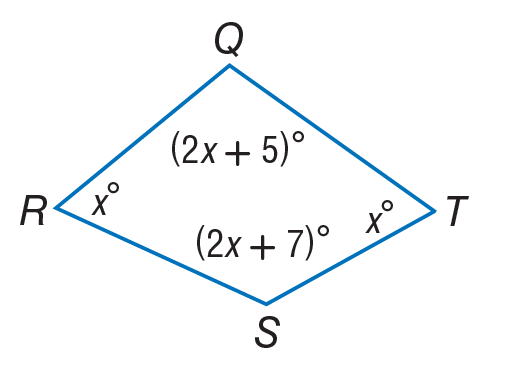}
\end{center}
\vspace{-10pt}
\caption{Diagram of a problem selected from Geometry3K.} 
\label{fig:geo3k_2489}
\end{figure}

The natural language version of the problem from the Geometry3K dataset is given below, and its corresponding diagram is shown in Figure~\ref{fig:geo3k_2489}. This example shows that our model can perform algebraic computation robustly after training on the synthetic dataset.
\begin{tcolorbox}[breakable, enhanced jigsaw]
\small{
\begin{lstlisting}[frame=none]
Find $m\angle R$.
\end{lstlisting}
}
\end{tcolorbox}

The solution produced by our model for this problem is shown below:
\begin{tcolorbox}[breakable, enhanced jigsaw]
\small{
\begin{lstlisting}[frame=none]
To find the measure of angle \( \angle R \), we need to use the fact that the sum of the interior angles of a quadrilateral is always 360 degrees.

Given the angles in the quadrilateral:
- \( \angle Q = (2x + 5)^\circ \)
- \( \angle R = x^\circ \)
- \( \angle S = (2x + 7)^\circ \)
- \( \angle T = x^\circ \)

We can set up the equation for the sum of the interior angles:
\[
(2x + 5) + x + (2x + 7) + x = 360
\]

Combine like terms:
\[
2x + 5 + x + 2x + 7 + x = 360
\]
\[
6x + 12 = 360
\]

Subtract 12 from both sides:
\[
6x = 348
\]

Divide by 6:
\[
x = 58
\]

Now, substitute \( x = 58 \) back into the expression for \( \angle R \):
\[
\angle R = x^\circ = 58^\circ
\]

Thus, the measure of \( \angle R \) is:
\[
\boxed{58}
\]
\end{lstlisting}
}
\end{tcolorbox}

The solution produced by the base model for this problem is shown below:
\begin{tcolorbox}[breakable, enhanced jigsaw]
\small{
\begin{lstlisting}[frame=none]
To find the measure of angle \( \angle R \) in the given quadrilateral, we need to use the fact that the sum of the interior angles of a quadrilateral is always \( 360^\circ \).

The given angles are:
- \( \angle Q = (2x + 5)^\circ \)
- \( \angle S = (2x + 7)^\circ \)
- \( \angle T = x^\circ \)
- \( \angle R = x^\circ \)

We can set up the equation for the sum of the interior angles:

\[
(2x + 5) + (2x + 7) + x + x = 360
\]

Combine like terms:

\[
2x + 5 + 2x + 7 + x + x = 360
\]

\[
4x + 12 = 360
\]

Subtract 12 from both sides:

\[
4x = 348
\]

Divide by 4:

\[
x = 87
\]

Now, substitute \( x = 87 \) back into the expression for \( \angle R \):

\[
\angle R = x^\circ = 87^\circ
\]

Thus, the measure of \( \angle R \) is:

\[
\boxed{87}
\]
\end{lstlisting}
}
\end{tcolorbox}

\subsection{Examples of Generated Formal Auxiliary Constructions}
\label{app:auxiliary_constructions}
We compared the auxiliary constructions identified by AlphaGeometry~\citep{alphageometry} with those generated by our approach. Notably, our predicted constructions often differ from those of AlphaGeometry, highlighting that multiple valid auxiliary strategies can achieve the same goal. Moreover, our method sometimes requires fewer auxiliary constructions than AlphaGeometry. To illustrate these differences, we present two examples from the IMO-AG-30 dataset~\citep{alphageometry}.

\begin{figure}[ht]
\begin{center}
\includegraphics[width=0.5\linewidth]{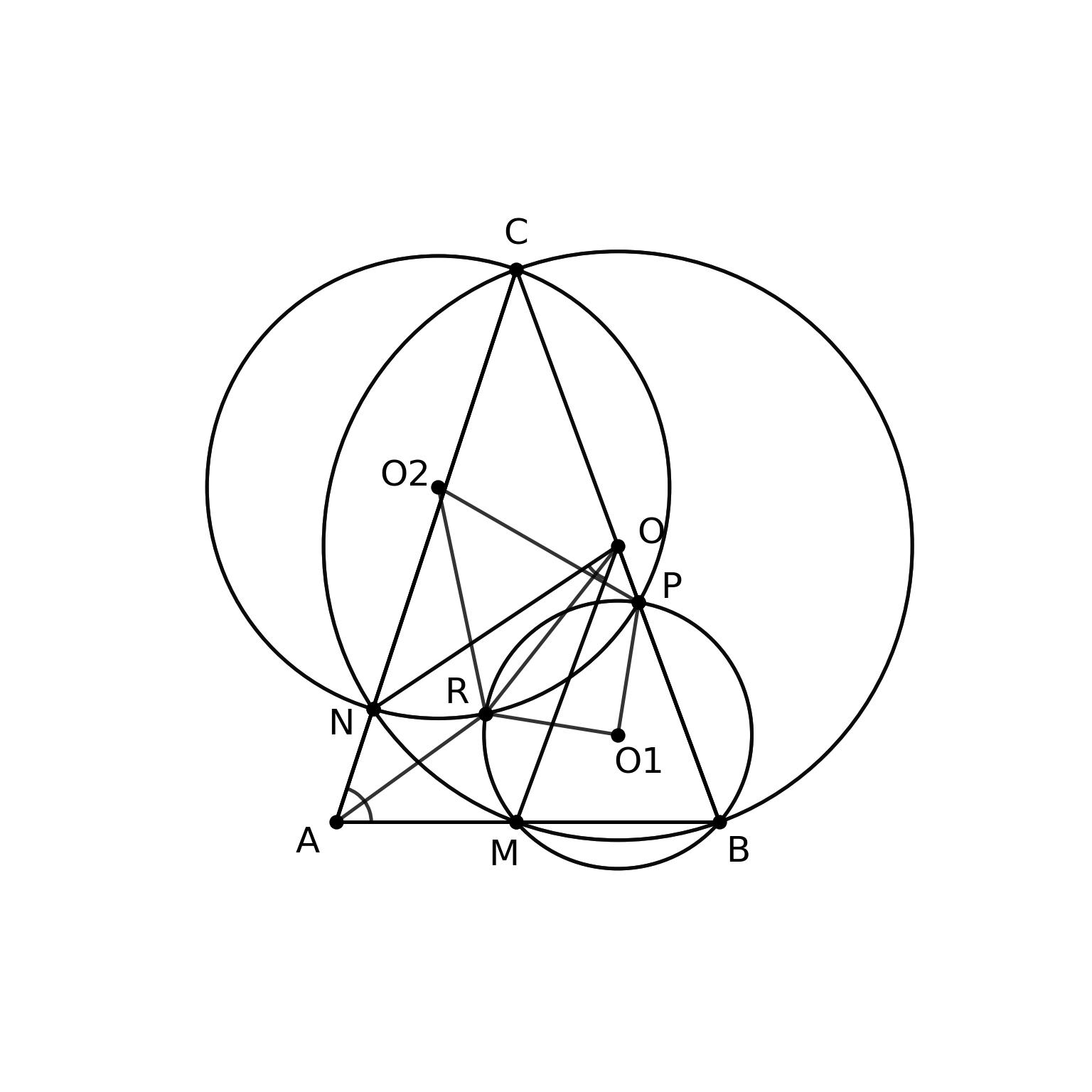}
\end{center}
\vspace{-10pt}
\caption{Diagram of a geometry problem selected from IMO-AG-30.} 
\label{fig:imo2004}
\end{figure}

The natural language formulation of the first problem is given below, and the corresponding diagram is shown in Figure~\ref{fig:imo2004}.
\begin{tcolorbox}[breakable, enhanced jigsaw]
\small{
\begin{lstlisting}[frame=none]
In triangle ABC, point O is the midpoint of side BC. Point M lies on the circle centered at O with radius OB, and M is located on line AB. Point N lies on the circle centered at O with radius OB, and N is located on line AC. Point R lies on the angle bisector of ∠BAC, such that ∠MOR = ∠RON. Let O_1 be the center of the circle passing through points B, M, and R, and O_2 be the circumcenter of triangle CNR. Point P lies on the circle centered at O_1 with radius O_1R, and P also lies on the circle centered at O_2 with radius O_2R. Prove that the points B, C, and P are collinear.
\end{lstlisting}
}
\end{tcolorbox}

The auxiliary constructions predicted by our LLM for this problem are shown below.
\begin{tcolorbox}[breakable, enhanced jigsaw]
\small{
\begin{lstlisting}[frame=none]
Construct point K as the circumcenter of triangle AMN.
\end{lstlisting}
}
\end{tcolorbox}

The auxiliary constructions predicted by AlphaGeometry for this problem are shown below.
\begin{tcolorbox}[breakable, enhanced jigsaw]
\small{
\begin{lstlisting}[frame=none]
Construct point K such that KM = KN.
Construct point L as the intersection of circles (K, A) and (O, A).
\end{lstlisting}
}
\end{tcolorbox}

\begin{figure}[ht]
\begin{center}
\includegraphics[width=0.5\linewidth]{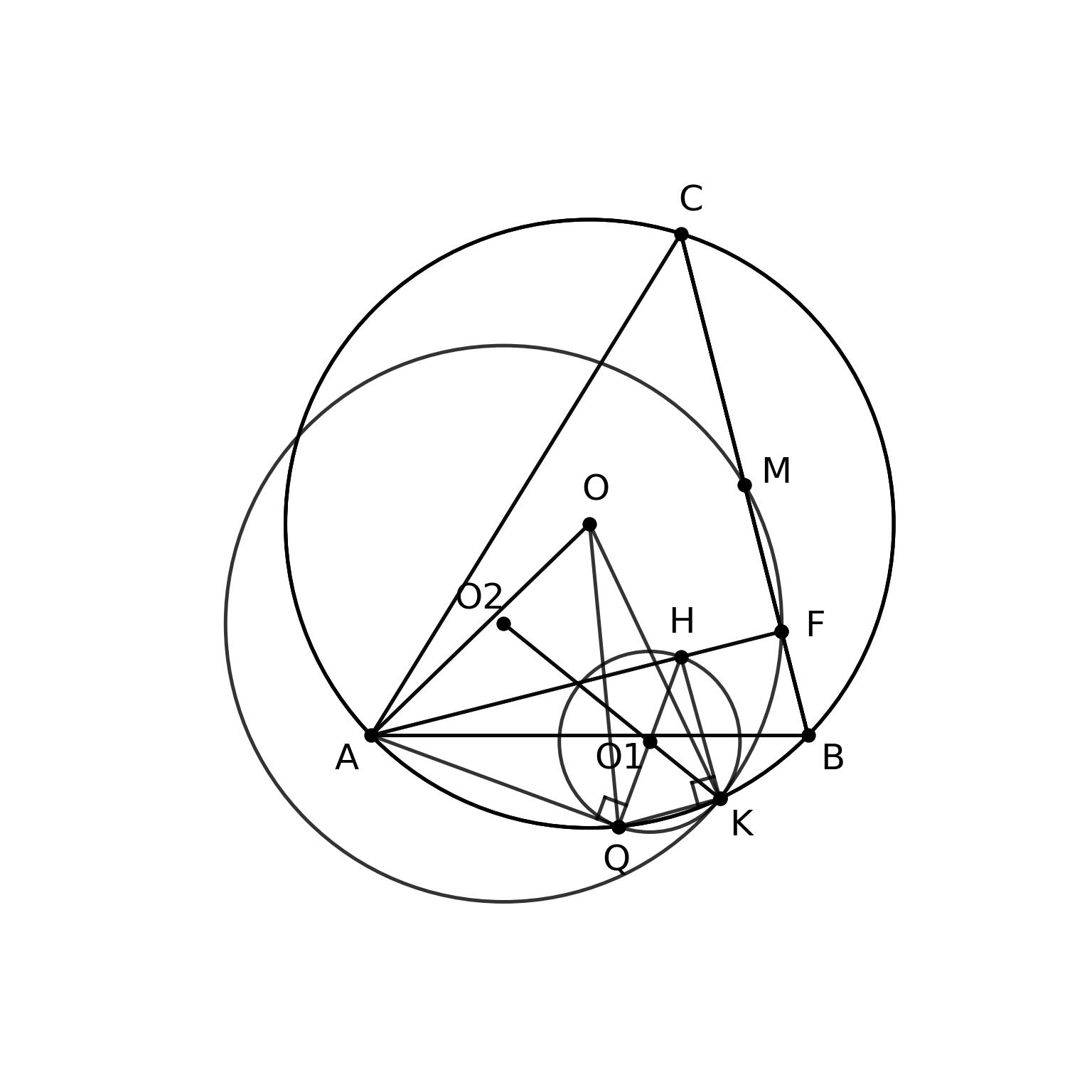}
\end{center}
\vspace{-10pt}
\caption{Diagram of a geometry problem selected from IMO-AG-30.} 
\label{fig:imo2015}
\end{figure}

The natural language formulation of the second problem is given below, and the corresponding diagram is shown in Figure~\ref{fig:imo2015}.
\begin{tcolorbox}[breakable, enhanced jigsaw]
\small{
\begin{lstlisting}[frame=none]
In triangle ABC, let H be the orthocenter. Point F lies on the line HA and also on the line BC. Let M be the midpoint of segment BC. Let O be the circumcenter of triangle ABC, which is the center of the circle passing through points A, B, and C. Triangle QAH is a right triangle with a 90-degree angle at Q, where Q lies on the circle centered at O with radius OA. Similarly, triangle KHQ is a right triangle with a 90-degree angle at K, where K also lies on the circle centered at O with radius OA. Let O_1 be the circumcenter of triangle KQH, and let O_2 be the circumcenter of triangle FKM. Prove that points K, O_1, and O_2 are collinear.
\end{lstlisting}
}
\end{tcolorbox}

The auxiliary constructions predicted by our LLM for this problem are shown below:
\begin{tcolorbox}[breakable, enhanced jigsaw]
\small{
\begin{lstlisting}[frame=none]
Construct point p as the intersection of cicle (O, A) and Line (H, Q).
\end{lstlisting}
}
\end{tcolorbox}

The auxiliary constructions predicted by AlphaGeometry for this problem are shown below:
\begin{tcolorbox}[breakable, enhanced jigsaw]
\small{
\begin{lstlisting}[frame=none]
Construct point X as the midpoint of CH.
Construct point Y as the midpoint of KM.
Construct point Z as the midpoint of BH.
\end{lstlisting}
}
\end{tcolorbox}


\end{document}